\documentclass[11pt]{article}
\usepackage[margin=1in]{geometry}
\usepackage{amsmath, amssymb, amsthm, mathtools}
\usepackage{booktabs, array, enumitem}
\usepackage{tikz}
\usetikzlibrary{positioning, calc, fit, shapes.geometric, decorations.pathreplacing}
\usepackage{graphicx}
\usepackage{subcaption}
\usepackage{xcolor}
\definecolor{linknavy}{rgb}{0.10,0.25,0.60}
\definecolor{linkgreen}{rgb}{0.10,0.45,0.25}
\definecolor{linkred}{rgb}{0.55,0.15,0.15}
\usepackage{amsthm}
\theoremstyle{plain}                     % bold label, italic body
\newtheorem{result}{Result}[section]     % numbered as Result 3.1, 3.2, ...
\usepackage[round]{natbib}               % author-year citations: \citet -> Name et al. (2019), \citep -> (Name et al., 2019)
\usepackage[colorlinks=true, linkcolor=linknavy, citecolor=linkgreen,
            urlcolor=linkred, breaklinks=true]{hyperref}

\newcommand{\D}{D}
\newcommand{\Eff}{\mathrm{eff}}
\newcommand{\modP}[1]{(#1)_{P}}
\newcommand{\Tr}{\mathrm{tr}\,}
\newcommand{\R}{\mathbb{R}}
\newcommand{\E}{\mathbb{E}}

\DeclareMathOperator{\erf}{erf}

\title{The role of locality and weight sharing\\ in convolutional neural networks}
\author{Francesco Mori$^{1}$ \qquad Melanie Weber$^{1,2}$ \\[6pt]
  {\normalsize $^{1}$ Center of Mathematical Sciences and Applications, Harvard University} \\
  {\normalsize $^{2}$ John A.\ Paulson School of Engineering and Applied Sciences, Harvard
  University}}
\date{\today}

\begin{document}
\maketitle

\begin{abstract}
The success of convolutional neural networks is often attributed to locality and weight sharing. However, how these two properties separately shape learning, and how their effects depend on the spatial correlations of natural images, remain poorly understood. We address these questions in a teacher--student model with a convolutional teacher acting on $P$ input positions with correlation $c$ between positions. We compare convolutional and locally connected students trained by online stochastic gradient descent, and derive ordinary differential equations that describe their dynamics exactly in the limit of large feature dimension $\D$.
For uncorrelated positions ($c = 0$), the two students follow the same learning trajectory, with the convolutional student learning exactly $P$ times faster. We also find that locality affects the sample complexity: a student that matches the kernel size of the teacher recovers it from $\mathcal{O}_{\D}(\D)$ samples, whereas $\mathcal{O}_{\D}(\D\log\D)$ samples are needed with a larger kernel. For $c > 0$, spatial correlations can reduce the advantage of weight sharing. The asymptotic error of the convolutional student grows with $c$, whereas that of the locally connected student is independent of $c$ and overall lower when evaluated at the same learning rate.
When the learning rate is instead tuned to the training horizon, however, the locally connected student wins at early times and the convolutional student at late times, with a transition that we predict analytically.
\end{abstract}

% =============================================================================
\section{Introduction}\label{sec:intro}
% =============================================================================

The ability of neural networks to learn from a limited number of examples depends on their
inductive biases, i.e., on the assumptions about the task that are built into the
architecture and the training algorithm. A natural source of inductive bias is symmetry. When
a task is invariant under a group of transformations of the input, a model that respects this
invariance does not need to learn the same features separately for every transformed version of
the input. Weight sharing is the standard way to build such symmetries into a
network \citep{ShaweTaylor1989, CohenWelling2016, RavanbakhshSchneiderPoczos2017,
KondorTrivedi2018}, and it is expected to reduce the number of samples needed to learn and to
speed up training.

The main success story of this approach is the convolutional neural network (CNN), widely
used in image classification and other computer vision tasks \citep{LeCun1998,
Krizhevsky2012}. CNNs exploit two structural properties of natural images, each encoded by a
distinct architectural choice: Weight sharing across positions encodes the approximate
invariance of image statistics under translations, while small filters encode the locality
of the relevant features. Additionally, natural images are strongly correlated across
space \citep{RudermanBialek1994}, and the labels used for training are often noisy, resulting
in a complex interplay between architectural constraints and the noise of stochastic
gradient descent (SGD).

From a theoretical perspective, building symmetries into a model is known to reduce the
number of samples needed to learn, typically by a factor of the order of the size of the
symmetry group \citep{BiettiVenturiBruna2021, MeiMisiakiewiczMontanari2021,
TahmasebiJegelka2023, ElesedyZaidi2021, LahotiKarpWinstonSinghLi2024, waber2026symmetry}.
However, disentangling the effects of locality and weight sharing remains a
largely open challenge, especially in the feature-learning regime. Moreover, most analyses
assume independent inputs, thereby neglecting the spatial correlations of natural images. It
is therefore unclear to what extent the advantage of weight sharing survives
in the presence of correlations and noise.

In this work, we address these questions in a teacher--student model whose learning dynamics
can be described exactly. The teacher is a two-layer convolutional network with kernel size
$k_t$, acting on inputs made of $P$ positions that each carry $\D$ features, with
correlation $c$ between positions. We train convolutional and locally connected students
with kernel size $k_s \geq k_t$ by online SGD. The two students have the same receptive
fields and differ only in weight sharing, while varying $k_s$ probes the role of locality.
In the limit of large feature dimension $\D$, the dynamics is described exactly by ordinary
differential equations (ODEs) for a few summary statistics, which extend the classical
analysis of online learning in two-layer networks \citep{SaadSolla1995, BiehlSchwarze1995,
GoldtAdvaniSaxeKrzakalaZdeborova2019}.

\paragraph{Contributions.} Our main contribution is an exact description of how weight sharing and locality shape the
online SGD dynamics of convolutional networks, which separates their effects. More precisely:
\begin{itemize}
\item In the limit of large feature dimension $\D$ and with Gaussian inputs, we derive analytically a closed set of ODEs for
the summary statistics of convolutional and locally connected students, for arbitrary kernel
sizes and input covariance
(Result~\ref{res:ODEs}).
\item For uncorrelated positions, we show analytically that the locally connected network follows
exactly the same training trajectory as the convolutional network, but needs exactly $P$ times
more SGD steps (Result~\ref{res:rescaling}). This is an
exact dynamical counterpart of the reduction in sample complexity by a factor of the order of
the group size found in previous works.
\item By integrating the ODEs numerically, we find that locality affects the sample complexity of
the problem. Indeed, a
convolutional student with the right degree of locality, $k_s = k_t$, learns the teacher
from $\mathcal{O}_{\D}(\D)$ samples, while $\mathcal{O}_{\D}(\D\log\D)$ samples are required
for $k_s > k_t$ (Result~\ref{res:locality}). The additional logarithmic factor originates
from a saddle point that is symmetric under reflection of the kernel, which the dynamics
escapes through the fluctuations of the random initialization.
\item We show that spatial correlations can reduce the advantage of weight sharing. With a
perturbative expansion at small learning rate, we find that the
asymptotic error of the convolutional network grows with the correlation
strength $c$, while that of the locally connected network is independent of $c$ and lower
than that of the convolutional network for any $c > 0$ when evaluated at the same learning rate (Result~\ref{res:floor}). However,
when the learning rate of each network is tuned to the training horizon, we find numerically
that the locally connected network has the lower error at early times but the convolutional
network wins at late times (Section~\ref{sec:phase}), with a transition time that we predict
analytically from an early-time expansion of the ODEs (Result~\ref{res:early}).
\end{itemize}

The code used in the paper is available at
\url{https://github.com/francescomori/CNN_theory}.

\paragraph{Outline.} The paper is organized as follows. Section~\ref{sec:setup} introduces the model.
Section~\ref{sec:OPs} presents the ODEs and the effects of weight sharing and locality for
uncorrelated positions. Section~\ref{sec:correlations} studies the effect of spatial
correlations, and Section~\ref{sec:conclusions} discusses limitations and future directions.
Related works are discussed in Section~\ref{sec:related_works}, and derivations and
additional results are given in the appendices.

% =============================================================================
\subsection{Related works}\label{sec:related_works}
% =============================================================================

\paragraph{Equivariant architectures.}
Weight sharing can be viewed as a mechanism for hard-coding symmetries into a learning
architecture. Early work constructed neural networks with tied parameters to guarantee
invariance under prescribed group actions \citep{ShaweTaylor1989}. This perspective later gave
rise to group-equivariant convolutional networks \citep{CohenWelling2016}, general
parameter-sharing constructions for equivariant layers \citep{RavanbakhshSchneiderPoczos2017},
and characterizations of convolution as the natural linear operation associated with
equivariance under group actions \citep{KondorTrivedi2018}.

\paragraph{Statistical benefits of symmetry.}
Another body of literature has quantified the statistical benefits of such symmetry
constraints. \citet{SannaiImaizumiKawano2021} derived generalization bounds for
invariant and equivariant networks using quotient feature spaces, while \citet{BiettiVenturiBruna2021} showed that learning with invariant kernels reduces the number
of samples by a factor asymptotically equal to the size of the symmetry group. Related sample-complexity gains were established for invariant random-feature and kernel
models \citep{MeiMisiakiewiczMontanari2021}, kernel regression \citep{TahmasebiJegelka2023}, and for equivariant
linear models \citep{ElesedyZaidi2021}. At the same time, symmetry alone does not remove computational
barriers: \citet{KianiLeLawrenceJegelkaWeber2024} established super-polynomial or
exponential statistical-query lower bounds for learning several classes of equivariant models,
including convolutional networks.

\paragraph{Locality versus weight sharing.}
Several works studied the separate contributions of locality and weight sharing.
In a teacher--student setting for kernel regression with convolutional kernels, \citet{FaveroCagnettaWyart2021} found that locality determines the exponent of the learning
curve, whereas translational invariance does not, and \citet{cagnetta2023can}
extended this analysis to deep convolutional kernels. \citet{WangWu2023} compared
convolutional, locally connected, and fully connected networks, and related the different
sample complexities of the three architectures to the symmetries of the learning procedure.
\citet{LahotiKarpWinstonSinghLi2024} compared the same three architectures on
inputs made of patches, with the signal planted in one of them, and found that the locally
connected network needs more samples than the convolutional one by a factor of order the
number of patches. Similarly, \citet{li2020convolutional} showed that, on some tasks, fully
connected networks trained by gradient descent need quadratically more samples than
convolutional ones. Regarding optimization, \citet{dAscoliSagunBrunaBiroli2019}
found experimentally that the convolutional constraint mainly helps during the early stages of
training, and \citet{IngrossoGoldt2022} showed that localized filters can
emerge in fully connected networks because of the non-Gaussian statistics of the inputs.

\paragraph{Learning convolutional teachers.}
For Gaussian inputs with independent, non-overlapping patches, it was shown that gradient
descent recovers a convolutional teacher in polynomial time \citep{globerson2017globally},
even when the loss has spurious local minima \citep{du2018gradient}.
\citet{goel2018learning} extended these results to overlapping patches and more general
input distributions, using a gradient-like algorithm. The very recent
work by \citet{waber2026symmetry} studied online SGD in a convolutional teacher--student model
with independent Gaussian inputs and a filter that extends over all patches, showing that weight sharing and data
augmentation reduce the sample complexity by a factor equal to
the number of patches, which is taken to be polynomial in the input dimension. In this work, we consider the roles of both weight sharing and locality, by considering a teacher
of arbitrary kernel size and comparing convolutional and locally connected students whose
kernel size can exceed that of the teacher. We also investigate how correlations across
patches can reduce the advantage of weight sharing, both in the speed of convergence and in
the asymptotic generalization error in the presence of label noise.

\paragraph{Exact learning dynamics in high dimensions.}
Classical works \citep{SaadSolla1995, BiehlSchwarze1995, EngelVandenBroeck2001} described the
dynamics of online learning in two-layer networks exactly in the high-dimensional limit,
through ordinary differential equations for a few order parameters. These equations were
used to explain the plateaus of the generalization error caused by the permutation symmetry
of the hidden units \citep{SaadSolla1995}, to predict the asymptotic error due to label
noise \citep{saad1997learning}, and to derive optimal learning-rate
schedules \citep{saad1997globally}. More recent works made this approach
rigorous \citep{GoldtAdvaniSaxeKrzakalaZdeborova2019, ben2022high} and extended it to
different learning regimes \citep{veiga2022phase}, structured
inputs \citep{goldt2020modeling}, and the design of optimal training
protocols \citep{mignacco2026statistical}. Here, we extend this framework to convolutional
and locally connected networks, which gives an exact description of their online SGD
dynamics. This allows us to quantify how locality sets the number of samples needed to
recover the teacher, when convolutional weight sharing accelerates learning relative to an
otherwise matched locally connected architecture, and when correlations and noise can
instead make the additional flexibility of the unshared model advantageous.

% =============================================================================
\section{Setup and notation}\label{sec:setup}
% =============================================================================

In this work, we study an exactly solvable teacher--student model in which the teacher is a convolutional network and the student is either convolutional or locally connected. Teacher--student models are a class of supervised learning settings in which the labels are generated by a fixed \emph{teacher} network, which represents the ground truth to be recovered, while a \emph{student}
network is trained to reproduce the teacher's output \citep{GardnerDerrida1989, EngelVandenBroeck2001}.
When the input distribution is sufficiently simple, for instance Gaussian, the learning dynamics of these models can often be characterized exactly in the high-dimensional limit \citep{SaadSolla1995, BiehlSchwarze1995, GoldtAdvaniSaxeKrzakalaZdeborova2019}.
Our convolutional teacher also belongs to the broader class of multi-index models \citep{arous2021online,abbe2023sgd,arnaboldi2024online,bietti2023learning,waber2026symmetry}.

\subsection{Data structure}

We consider a supervised learning setting with inputs $x$ and labels $y$. The inputs are vectors $x \in \R^{P \D}$ organized as
\begin{equation}
    x \;=\; (x_0, \ldots, x_{P-1}), \qquad x_p \in \R^{\D}.
\end{equation}
We refer to the index $p \in \{0, \ldots, P - 1\}$ as the \emph{position} and to the $\D$
entries of each $x_p$ as the \emph{features} at that position.  We
take $P$ fixed and $\D \to \infty$. This regime is natural for the deeper layers of a
convolutional network, where each spatial position carries a high-dimensional feature
(many channels) while the spatial extent is comparatively small.

The components of $x$ are jointly Gaussian with mean zero and covariance
\begin{equation}
    \E\big[x_{p, a}\, x_{q, b}\big] \;=\; S_{p, q}\,\delta_{a, b}, \qquad p, q \in \{0, \ldots, P - 1\},\quad a, b \in \{1, \ldots, \D\}.
\end{equation}
The $P \times P$ matrix $S$ captures the spatial structure of the data. We will often assume the compound-symmetric form
\begin{equation}\label{eq:S-compound}
    S_{p, q} \;=\;
    \begin{cases}
        1, & p = q, \\
        c, & p \neq q,
    \end{cases}
    \qquad c \in [0, 1),
\end{equation}
so that every position has unit variance and every pair of distinct positions carries the
same correlation $c$. At $c =0$ it reduces to the case of independent positions $S = I$.

\begin{figure}[t]
\centering
\resizebox{\textwidth}{!}{%
\begin{tikzpicture}[
    patch/.style={draw, rectangle, minimum width=0.9cm, minimum height=0.45cm, fill=blue!8, font=\small, inner sep=2pt},
    unit/.style={draw, circle, minimum size=0.5cm, inner sep=0pt, font=\scriptsize},
    sumop/.style={draw, circle, minimum size=0.5cm, inner sep=0pt, font=\small, thick},
    teach/.style={draw=orange!85!black, line width=0.7pt, opacity=0.9},
    student/.style={draw=blue!75!black, line width=0.7pt, opacity=0.9},
    Bclr0/.style={draw=red!80!black, line width=0.9pt},
    Bclr1/.style={draw=blue!70!black, line width=0.9pt},
    Bclr2/.style={draw=green!55!black, line width=0.9pt},
    Bclr3/.style={draw=orange!90!black, line width=0.9pt},
    every node/.style={font=\small},
    >=stealth,
    panelwidth/.style={minimum width=4cm},
]

% ====== Panel: Teacher (left) ======
\node[anchor=south, align=center, font=\bfseries] at (1.6, 2.85) {Teacher};
\foreach \j in {0, 1, 2, 3} {
    \node[patch] (xT\j) at (1.0*\j + 0.2, 2.2) {$x_{\j}$};
}
\foreach \j/\val in {0/0, 1/1, 2/2, 3/3} {
    \node[unit] (lT\j) at (1.0*\j + 0.2, 0.9) {$\lambda_{\val}$};
}
% Connections: lambda_i <- x_p (w_0), x_{i+1} (w_1).
\foreach \j in {0, 1, 2} {
    \pgfmathtruncatemacro{\jn}{\j+1}
    \draw[teach] (xT\j.south) -- (lT\j.north);
    \draw[teach] (xT\jn.south) -- (lT\j.north);
}
% lambda_3 cyclic wrap: from x_3 and x_0 (drawn going below)
\draw[teach] (xT3.south) -- (lT3.north);
\draw[teach, rounded corners=4pt] (xT0.north) -- (0.2, 2.72) -- (4.05, 2.72) -- (4.05, 0.9) -- (lT3.east);
% Sum
\node[sumop] (sT) at (1.7, -0.3) {$\Sigma$};
\foreach \j in {0, 1, 2, 3} { \draw[->, black, line width=0.7pt] (lT\j) -- (sT); }
\node[right=3pt of sT, font=\footnotesize] {$+\,\sigma\xi = y$};

% ====== Panel: CNN (center) ======
\begin{scope}[xshift=7cm]
\node[anchor=south, align=center, font=\bfseries] at (1.6, 2.85) {CNN};
\foreach \j in {0, 1, 2, 3} {
    \node[patch] (xA\j) at (1.0*\j + 0.2, 2.2) {$x_{\j}$};
}
\foreach \j/\val in {0/0, 1/1, 2/2, 3/3} {
    \node[unit] (lA\j) at (1.0*\j + 0.2, 0.9) {$\hat\lambda_{\val}$};
}
% hat-lambda_i <- x_p, x_{i+1}, x_{i+2}: in-range taps
\draw[student] (xA0.south) -- (lA0.north);
\draw[student] (xA1.south) -- (lA0.north);
\draw[student] (xA2.south) -- (lA0.north);
\draw[student] (xA1.south) -- (lA1.north);
\draw[student] (xA2.south) -- (lA1.north);
\draw[student] (xA3.south) -- (lA1.north);
\draw[student] (xA2.south) -- (lA2.north);
\draw[student] (xA3.south) -- (lA2.north);
\draw[student] (xA3.south) -- (lA3.north);
% cyclic wraps, routed around the outside
\draw[student, rounded corners=4pt] (xA0.north) -- (0.2, 2.72) -- (4.05, 2.72) -- (4.05, 0.35) -- (2.2, 0.35) -- (lA2.south);
\draw[student, rounded corners=4pt] (xA1.north) -- (1.2, 2.52) -- (3.78, 2.52) -- (3.78, 0.9) -- (lA3.east);
\draw[student, rounded corners=4pt] (xA0.west) -- (-0.55, 2.2) -- (-0.55, -0.95) -- (4.3, -0.95) -- (4.3, 0.72) -- (lA3.south east);
\node[sumop] (sA) at (1.7, -0.3) {$\Sigma$};
\foreach \j in {0, 1, 2, 3} { \draw[->, black, line width=0.7pt] (lA\j) -- (sA); }
\node[right=3pt of sA, font=\footnotesize] {$= \hat y_{\mathrm{CNN}}$};
\end{scope}

% ====== Panel: LCN (right) ======
\begin{scope}[xshift=14cm]
\node[anchor=south, align=center, font=\bfseries] at (1.6, 2.85) {LCN};
\foreach \j in {0, 1, 2, 3} {
    \node[patch] (xB\j) at (1.0*\j + 0.2, 2.2) {$x_{\j}$};
}
\foreach \j/\val in {0/0, 1/1, 2/2, 3/3} {
    \node[unit] (lB\j) at (1.0*\j + 0.2, 0.9) {$\hat\lambda_{\val}$};
}
% Different color per output unit (per-position kernel), kernel size 3
\draw[Bclr0] (xB0.south) -- (lB0.north);
\draw[Bclr0] (xB1.south) -- (lB0.north);
\draw[Bclr0] (xB2.south) -- (lB0.north);
\draw[Bclr1] (xB1.south) -- (lB1.north);
\draw[Bclr1] (xB2.south) -- (lB1.north);
\draw[Bclr1] (xB3.south) -- (lB1.north);
\draw[Bclr2] (xB2.south) -- (lB2.north);
\draw[Bclr2] (xB3.south) -- (lB2.north);
\draw[Bclr3] (xB3.south) -- (lB3.north);
% cyclic wraps, routed around the outside
\draw[Bclr2, rounded corners=4pt] (xB0.north) -- (0.2, 2.72) -- (4.05, 2.72) -- (4.05, 0.35) -- (2.2, 0.35) -- (lB2.south);
\draw[Bclr3, rounded corners=4pt] (xB1.north) -- (1.2, 2.52) -- (3.78, 2.52) -- (3.78, 0.9) -- (lB3.east);
\draw[Bclr3, rounded corners=4pt] (xB0.west) -- (-0.55, 2.2) -- (-0.55, -0.95) -- (4.3, -0.95) -- (4.3, 0.72) -- (lB3.south east);
\node[sumop] (sB) at (1.7, -0.3) {$\Sigma$};
\foreach \j in {0, 1, 2, 3} { \draw[->, black, line width=0.7pt] (lB\j) -- (sB); }
\node[right=3pt of sB, font=\footnotesize] {$= \hat y_{\mathrm{LCN}}$};
\end{scope}

\end{tikzpicture}%
}
\caption{Teacher and the two student architectures with $P$ input positions. Colors indicate weight sharing. The
teacher is a two-layer CNN with kernel size $k_t$. The CNN student has kernel size $k_s$ and weight sharing. The LCN has the same receptive fields as the CNN but no weight sharing. Boundary conditions are cyclic. Here, $P=4$, $k_s=3$, and $k_t=2$.}
\label{fig:archs}
\end{figure}
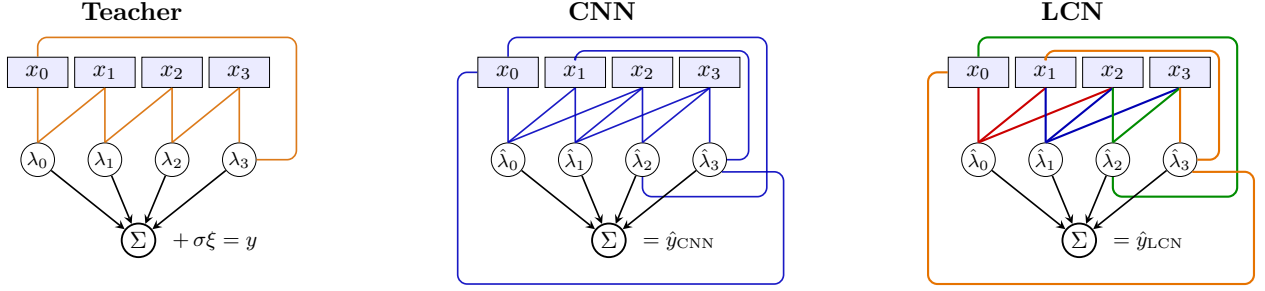

The labels are generated as 
\begin{equation}
    y(x) = f^\star(x) + \sigma\,\xi  \,,
\end{equation}
where $f^\star(x)$ is the output of the teacher network, $\xi\sim \mathcal{N}(0,1)$ independent of $x$, and $\sigma\geq 0$ is the standard deviation of the label noise. The teacher is a single-hidden-unit convolutional network with kernel size $k_t < P$ and
cyclic boundary conditions
\begin{equation}\label{eq:teacher-output}
    f^\star(x) \;=\; \sum_{p = 0}^{P - 1} g\big(\lambda_p(x)\big),
\end{equation}
with the preactivations 
\begin{equation}\label{eq:teacher-pre-act}
    \lambda_p(x) \;=\; \frac{1}{\sqrt{k_t\D}}\sum_{l = 0}^{k_t - 1} w_l \cdot x_{\modP{p + l}},
\end{equation}
and $\modP{q} \equiv q \bmod P$. The weight vectors $w_0, \ldots, w_{k_t - 1}$ form the
\emph{kernel} (or \emph{filter}) of the teacher, which is applied at every position $p$
to the $k_t$ consecutive positions $p, \ldots, p + k_t - 1$, and $k_t$ is the kernel size. Here, $g(z)$ indicates a nonlinear function. For analytic convenience, 
we will focus on $g(z) = \erf(z / \sqrt 2)$. We consider the case of ReLU activation numerically in
Appendix~\ref{app:relu}. We take the teacher's weight vectors to be
mutually orthogonal and of equal norm,
\begin{equation}\label{eq:teacher-ortho}
    \frac{w_l \cdot w_{l'}}{\D} \;=\; T\,\delta_{l, l'}\,.
\end{equation}

\subsection{Student architectures}

As students, we consider both convolutional neural networks (CNNs) and locally connected networks (LCNs). 
In both cases we take the kernel size $k_s\geq k_t$, so that the student can always represent the teacher 
but could in principle have a larger receptive field. 

The CNN student takes the form
\begin{equation}
    \hat y_{\mathrm{CNN}}(x) = \sum_{p = 0}^{P - 1} g(\hat\lambda_p^{\mathrm{CNN}}),
\end{equation}
with preactivations
\begin{equation}
    \hat\lambda_p^{\mathrm{CNN}}(x) \;=\; \frac{1}{\sqrt{k_s\D}}\sum_{l = 0}^{k_s - 1} \hat w_l \cdot x_{\modP{p + l}}.
\end{equation}
The same kernel $(\hat w_l)_{l = 0}^{k_s - 1}$ is used at every position, implementing translational invariance. 
Note that at $k_s=k_t$ the CNN student exactly matches the teacher architecture.
The LCN student is given by
\begin{equation}
    \hat y_{\mathrm{LCN}}(x) = \sum_{p = 0}^{P - 1} g(\hat\lambda_p^{\mathrm{LCN}}),
\end{equation}
with preactivations
\begin{equation}
    \hat\lambda_p^{\mathrm{LCN}}(x) \;=\; \frac{1}{\sqrt{k_s\D}}\sum_{l = 0}^{k_s - 1} \hat w^{(p)}_l \cdot x_{\modP{p + l}}.
\end{equation}
Each position $p$ has its own kernel $(\hat w^{(p)}_l)_{l = 0}^{k_s - 1}$, so the LCN has $P$ times as many trainable parameters as the CNN.
Note that at $k_s = P$ each unit of the LCN reads the whole input, resulting in a soft-committee machine \citep{SaadSolla1995}.
A schematic of the two architectures is given in Figure~\ref{fig:archs}.

%\mw{It may be instructive to add an example for how this structure implements symmetries as inductive bias. Translations on an image "pixel grid" with periodic boundary conditions. Is the framework specific to translation-invariant architectures or are, e.g., permutation-invariant settings covered?}

Weight sharing and locality are the two structural properties that
distinguish CNNs. Interpreting the input as a one-dimensional image with
periodic boundary conditions, weight sharing makes both the teacher and the
CNN student invariant under translations. Indeed, defining the cyclic shift
$(\tau x)_p = x_{(p+1)_P}$, one can easily check that $f^{\star}(\tau x)=f^{\star}(x)$
and $\hat{y}_{\rm CNN}(\tau x)=\hat{y}_{\rm CNN}(x)$. When $S$ is compound symmetric,
Eq.~\eqref{eq:S-compound}, the input distribution is also invariant under $\tau$, and the CNN
student builds the symmetry
of the task into its architecture. The LCN student, in contrast, is in general
invariant only when all its kernels coincide, $\hat w^{(p)}_l = \hat w_l$, and
must therefore learn the symmetry from data. % The first question we address is how this inductive bias affects the learning dynamics. We also study the role of locality by considering students with kernel size $k_s \geq k_t$. As we will show, matching the receptive field of the student to that of the teacher, $k_s = k_t$, reduces the number of samples needed to recover the teacher from $\mathcal{O}_{\D}(D\log D)$ to $\mathcal{O}_{\D}(D)$.

\subsection{Learning algorithm}\label{sec:setup-sgd}

Both students are trained by online stochastic gradient descent (SGD) on the squared loss
\begin{equation}
    \ell(x; \theta) \;=\; \tfrac{1}{2}\big(y(x) - \hat y_\theta(x)\big)^2,
\end{equation}
where $\theta$ collectively denotes the trainable weights of the student. 
At each SGD step $t$ a fresh sample $(x^t, y^t)$ is drawn from the data distribution,
independently of all previous ones, and the weights are updated by
\begin{equation}\label{eq:sgd-update}
    \theta^{t + 1} \;=\; \theta^{t} \;-\; \eta\,\nabla_\theta\,\ell\big(x^t; \theta^t\big),
\end{equation}
with $\eta > 0$ the learning rate. We initialize the student at random, drawing every component of its weights independently from $\mathcal{N}(0, 1)$. We measure performance by the generalization error
\begin{equation}
    \varepsilon_g(\theta) \;=\; \E_x\Big[\big(f^\star(x) - \hat y_\theta(x)\big)^2\Big].
\end{equation}
As we show in Appendix~\ref{app:gen-error}, $\varepsilon_g$ depends on
$\theta$ only through low-dimensional summary statistics.

% =============================================================================
\section{Learning dynamics}\label{sec:OPs}
% =============================================================================

In the limit where the feature dimension $\D$ and the number of SGD steps $t$
jointly tend to infinity at a fixed ratio $ \alpha \;\equiv\; t / \D,$ the network's dynamics is captured by a small set of low-dimensional summary statistics,
known as \emph{order parameters} in statistical physics. As stated in the Result below, the order parameters
concentrate as $\D \to \infty$ and satisfy a closed system of ordinary differential
equations (ODEs) in $\alpha$. The CNN student is described by two matrices of order
parameters,
\begin{equation}\label{eq:OP_CNN}
    Q_{l l'} \;=\; \frac{\hat w_l \cdot \hat w_{l'}}{k_s\D}, \qquad
    R_{l l''} \;=\; \frac{\hat w_l \cdot w_{l''}}{\sqrt{k_sk_t}\,\D}\,,
\end{equation}
with $l, l' \in \{0, \ldots, k_s - 1\}$ and $l'' \in \{0, \ldots, k_t - 1\}$. The LCN student requires two additional position indices $p, q \in \{0, \ldots, P - 1\}$,
\begin{equation}\label{eq:OP_LCN}
    Q^{(p, q)}_{l l'} \;=\; \frac{\hat w^{(p)}_l \cdot \hat w^{(q)}_{l'}}{k_s\D}, \qquad
    R^{(p)}_{l l''} \;=\; \frac{\hat w^{(p)}_l \cdot w_{l''}}{\sqrt{k_sk_t}\,\D}\,.
\end{equation}

\begin{result}[Exact ODEs for the learning dynamics]\label{res:ODEs}
Consider the limit $\D, t \to \infty$ with $\alpha = t/\D = \mathcal{O}_{\D}(1)$, with
$P$, $k_t$, $k_s$, $\eta$, $T$, and $\sigma$ held fixed.
\begin{itemize}
\item[(i)] \emph{CNN student.} Let $\hat w^t_l$
denote the weights of the CNN after $t$ steps of online
SGD~\eqref{eq:sgd-update}. The associated order parameters $R$ and $Q$, defined in Eq.~\eqref{eq:OP_CNN}, converge in
probability to the solution at time $\alpha$ of the deterministic ODEs
\begin{equation}\label{eq:ODEs-CNN}
    \frac{dR}{d\alpha} = \frac{\eta}{k_s}\,F^{\rm CNN}_R(R, Q), \qquad
    \frac{dQ}{d\alpha} = \frac{\eta}{k_s}\,F^{\rm CNN}_Q(R, Q)
    + \frac{\eta^2}{k_s^2}\,G^{\rm CNN}_Q(R, Q).
\end{equation}
\item[(ii)] \emph{LCN student.} Let
$\hat w^{(p),t}_l$ denote the weights of the LCN after
$t$ steps of online SGD. The associated order parameters $R^{(p)}$ and $Q^{(p,q)}$, defined in Eq.~\eqref{eq:OP_LCN},
converge in probability to the solution at time $\alpha$ of
\begin{equation}\label{eq:ODEs-LCN}
    \frac{dR^{(p)}}{d\alpha} = \frac{\eta}{k_s}\,F^{{\rm LCN},(p)}_R(\mathbf R, \mathbf Q), \qquad
    \frac{dQ^{(p,q)}}{d\alpha} = \frac{\eta}{k_s}\,F^{{\rm LCN},(p,q)}_Q(\mathbf R, \mathbf Q)
    + \frac{\eta^2}{k_s^2}\,G^{{\rm LCN},(p,q)}_Q(\mathbf R, \mathbf Q),
\end{equation}
where $\mathbf R = \{R^{(p)}\}_p$ and $\mathbf Q = \{Q^{(p,q)}\}_{p,q}$.
\end{itemize}
The functions $F_R$, $F_Q$, and $G_Q$ are derived in
Appendices~\ref{app:ode-A} and~\ref{app:ode-B}.
\end{result}

The terms linear in $\eta$ correspond to gradient flow on the population
loss, while the term proportional to $\eta^2$ describes SGD noise. We notice that while the theory
is in principle only valid in the thermodynamic limit $\D\to \infty$, it matches
numerical simulations already at $\D = 1000$, as shown in Figure~\ref{fig:rescaling}. %The ODEs \eqref{eq:ODEs-CNN} and \eqref{eq:ODEs-LCN} are our first main theoretical result and will be used for more detailed analysis in the rest of the paper. In the rest of this section, we investigate how translational invariance and locality influence the training dynamics.

Note that since each weight component is initialized as an independent standard Gaussian variable, the initial conditions for large $\D$ are
\begin{equation}\label{eq:init-OP-CNN}
    Q_{l l'} = \frac{\delta_{l l'}}{k_s} + \mathcal{O}_{\D}(\D^{-1/2})\,, \qquad
    R_{l l'} = \mathcal{O}_{\D}(\D^{-1/2})
\end{equation}
for the CNN, and
\begin{equation}\label{eq:init-OP-LCN}
    Q^{(p, q)}_{l l'} = \delta_{p q}\,\delta_{l l'}/k_s + \mathcal{O}_{\D}(\D^{-1/2})\,, \qquad
    R^{(p)}_{l l'} = \mathcal{O}_{\D}(\D^{-1/2})\,,
\end{equation}
for the LCN.

\subsection{Benefits of translational invariance}\label{sec:AvsB-rescaling}

\begin{figure}[t]
    \centering
    \begin{subfigure}{0.48\textwidth}\centering
      \includegraphics[width=\textwidth]{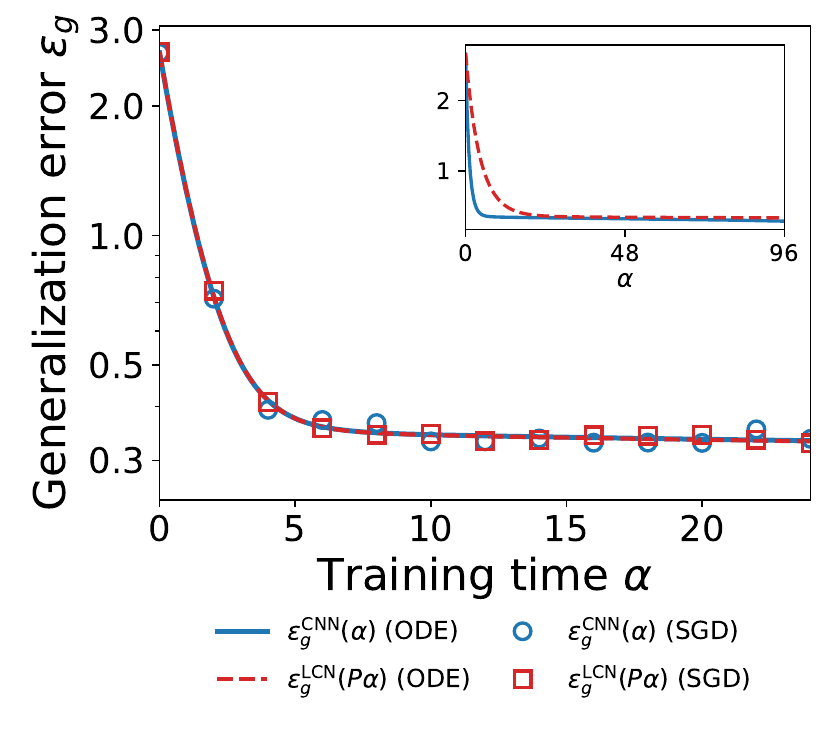}
      \caption{}\label{fig:rescaling-a}
    \end{subfigure}
    \hfill
    \begin{subfigure}{0.48\textwidth}\centering
      \includegraphics[width=\textwidth]{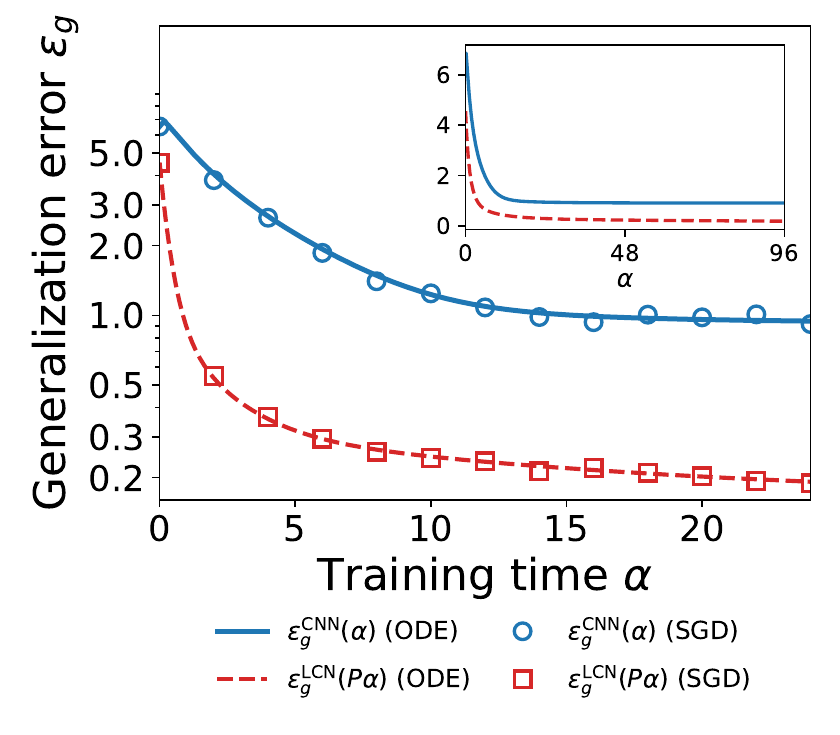}
      \caption{}\label{fig:rescaling-b}
    \end{subfigure}
    \caption{Generalization errors $\varepsilon_g^{\rm CNN}(\alpha)$ of the CNN
    and $\varepsilon_g^{\rm LCN}(P\alpha)$ of the LCN.
    Lines are the order-parameter ODEs and symbols mark simulations of a single SGD trajectory. Insets show the ODE curves without the time rescaling. (a) For independent positions ($c=0$) the trajectories coincide. (b) In the presence of spatial correlations ($c = 0.5$) the rescaling fails and the LCN converges faster (see inset).
    Parameters: $g = \erf$, $k_t = 2$, $k_s = 3$, $P = 4$, $\eta
    = 1$, $\sigma = 0$, $T = 1$, SGD at $\D = 1000$.}
    \label{fig:rescaling}
    \end{figure}

We next compare the dynamics of the CNN and the LCN, starting from the simplest case of uncorrelated positions, $S = I$. In this case, weight sharing has a remarkably simple effect: it speeds up the dynamics by exactly a factor $P$.

\begin{result}[Exact $P$-fold speedup from weight sharing]\label{res:rescaling}
Let $S = I$ and $k_t \le k_s \le P$. Let $(R, Q)(\alpha)$ be the solution of
the CNN equations~\eqref{eq:ODEs-CNN} with initial condition $(R_0, Q_0)$,
and let $(\mathbf R, \mathbf Q)(\alpha)$ be the solution of the LCN
equations~\eqref{eq:ODEs-LCN} with the matched, position-symmetric initial
condition
\begin{equation}
    R^{(p)}(0) = R_0, \qquad
    Q^{(p,\modP{p+l-l'})}_{ll'}(0) = (Q_0)_{ll'} \qquad \text{for all } p.
\end{equation}
Then, for all $\alpha \ge 0$ and all $p$,
\begin{equation}\label{eq:rescaling}
    R^{(p)}(P\alpha) = R(\alpha), \qquad
    Q^{(p,\modP{p+l-l'})}_{ll'}(P\alpha) = Q_{ll'}(\alpha), \qquad
    \varepsilon_g^{\rm LCN}(P\alpha) = \varepsilon_g^{\rm CNN}(\alpha).
\end{equation} This result can be obtained by direct inspection of the ODEs~\eqref{eq:ODEs-CNN} and \eqref{eq:ODEs-LCN} (see
Appendix~\ref{app:rescaling} for details). Note that independent Gaussian
initialization satisfies the position-symmetry condition up to fluctuations of order $1/\sqrt{\D}$ (see Eq.~\eqref{eq:init-OP-LCN}).
\end{result}

Figure~\ref{fig:rescaling-a} illustrates this result. As shown in the inset, the generalization error of the CNN decreases faster than that of the LCN. After rescaling the time of the LCN by a factor $P$, the two error curves collapse onto each other (main panel), both for the ODEs and for the SGD simulations. The origin of the factor $P$ can be understood as follows. At each sample, the shared kernel of the CNN accumulates a gradient contribution from every one of the $P$ positions, while each of the $P$ kernels of the LCN only receives the
contribution from its own position. Quite surprisingly, this rescaling is exact even at finite learning rate $\eta$, when the effects of the SGD noise are not negligible.

This result aligns with several previous works showing that invariant models can have a lower sample complexity than their non-invariant counterparts, by a factor of the order of the size of the associated symmetry group (here the cyclic group of translations of size $P$). This was shown for convolutional networks trained with gradient-based algorithms \citep{LahotiKarpWinstonSinghLi2024, waber2026symmetry}, and in more general settings for invariant classifiers \citep{sokolic2017generalization} and for invariant kernel and random-feature methods \citep{BiettiVenturiBruna2021, MeiMisiakiewiczMontanari2021,
TahmasebiJegelka2023}. In contrast with these results, which take the form of bounds or asymptotic rates, Result~\ref{res:rescaling} establishes an exact identity between the full training trajectories of the two students.

Interestingly, the exact rescaling holds only for uncorrelated positions. As shown in Figure~\ref{fig:rescaling-b}, for $c > 0$ the CNN at time $\alpha$ and the LCN at time $P\alpha$ follow visibly different curves. Strikingly, even without rescaling time, the LCN converges faster than the CNN. This suggests that a less constrained architecture can better adapt to correlations across positions, a point we make precise in the next sections.
% =============================================================================
\subsection{Benefits of locality}\label{sec:speed}
% =============================================================================

%\mw{Early on in the section I would summarize cleanly in one sentence what the impact of weight sharing is.}

\begin{figure}[t]
\centering
\begin{subfigure}{0.48\textwidth}\centering
  \includegraphics[width=\textwidth]{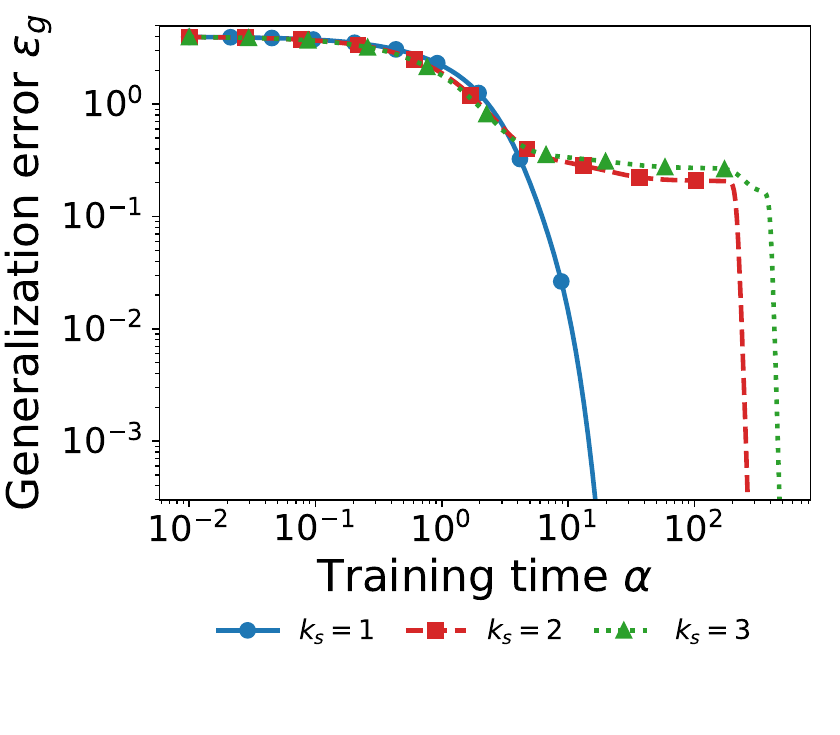}
  \caption{}\label{fig:plateau-a}
\end{subfigure}
\hfill
\begin{subfigure}{0.48\textwidth}\centering
  \includegraphics[width=\textwidth]{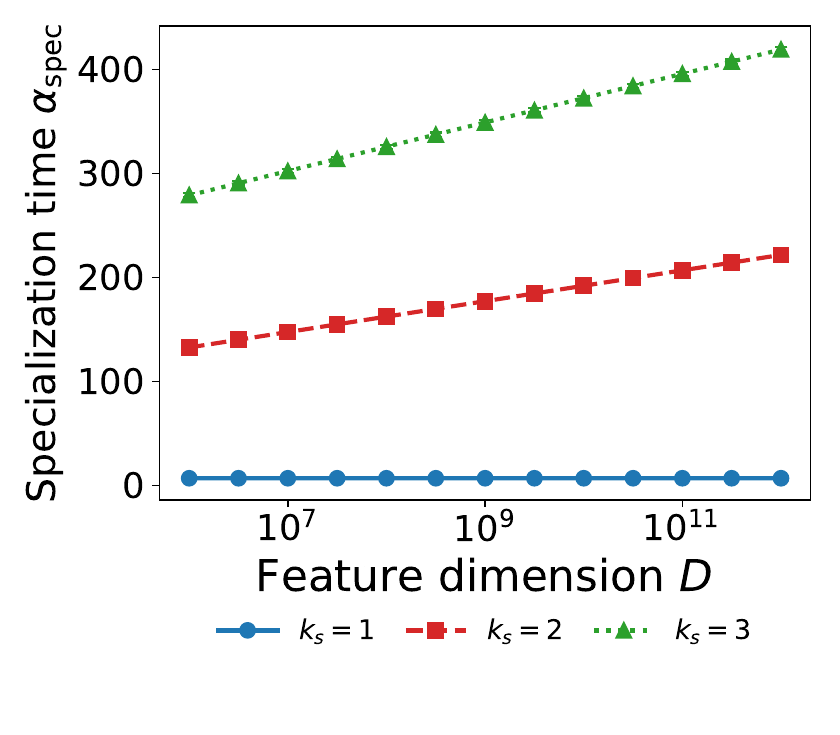}
  \caption{}\label{fig:plateau-b}
\end{subfigure}
\caption{CNN students that match the teacher kernel size ($k_s=k_t$) learn with $\mathcal{O}_{\D}(\D)$ samples,
while students with $k_s>k_t$ take $\mathcal{O}_{\D}(\D\log\D)$ samples. (a) Generalization error
vs.~training time $\alpha$ for three kernel sizes, obtained by integrating the
ODEs~\eqref{eq:ODEs-CNN} from the order parameters of a random initialization at dimension $\D$.
(b) Specialization time $\alpha_{\rm spec}$ (Eq.~\eqref{eq:alpha-spec}, with
$\gamma_{\rm th} = 0.99$) against the feature dimension $\D$. Markers are averages over random
initializations, each obtained by integrating the ODEs~\eqref{eq:ODEs-CNN} from the order
parameters of a random initialization at dimension $\D$. Standard errors are smaller than the marker size. Lines are fits $a + b\ln\D$. Parameters: $\D = 10^{12}$ in panel (a),
$g = \erf$, $k_t = 1$, $P = 6$, $S = I$, $\eta = 0.5$, $T = 1$, $\sigma = 0$, 128 random
initial conditions in panel (b).}
\label{fig:plateau}
\end{figure}

In this section, we study how locality affects the learning dynamics. For simplicity, we focus on the CNN student and investigate the role of the kernel size $k_s$, which determines the degree of locality of the network. By integrating the ODEs numerically, we find that a student that matches the teacher locality ($k_s=k_t$) requires $\mathcal{O}_D(D)$ samples to learn the teacher. On the other hand, for $k_s>k_t$, $\mathcal{O}_D(D\log D)$ samples are required.

It is instructive to first enumerate the student configurations that exactly reproduce the teacher output, achieving zero generalization error. For $k_s,k_t<P$, there are $k_s - k_t + 1$ solutions
\begin{equation}\label{eq:degenerate}
    \hat w_l \;=\; \sqrt{k_s/k_t}\;\, w_{l - s} \quad\text{for}\quad s \le l < s+k_t,
    \qquad \hat w_l \;=\; 0 \quad\text{otherwise}\,,
\end{equation}
labeled by the shift $s \in \{0, \ldots, k_s - k_t\}$. Note that for $k_s=k_t<P$ the solution is unique. We define the cosine similarity of the student parameters with the $s$-th solution as 
\begin{equation}\label{eq:gamma}
    \gamma_s \;=\; \frac{1}{\sqrt{T\,\Tr Q}}\sum_{l = 0}^{k_t - 1} R_{s + l,\, l}\,.
\end{equation}
Note that when the student specializes to a particular solution $s$, one has $\gamma_s=1$ and $\gamma_{s'}=0$ for $s'\neq s$. As shown below, although the ODEs~\eqref{eq:ODEs-CNN} do not depend explicitly on $\D$, the training time $\alpha_{\rm spec}$ at which specialization occurs can depend on $D$ through the random initial conditions in Eq.~\eqref{eq:init-OP-CNN}.

\begin{result}[Locality and specialization time, numerical]\label{res:locality}
Let $S = I$ and $\sigma = 0$, let $\eta$ be small enough to ensure convergence to a zero-error
solution, and let $(R, Q)(\alpha)$ be the solution of the
ODEs~\eqref{eq:ODEs-CNN} from random initial conditions
in dimension $\D$. For a fixed threshold $\gamma_{\rm th} < 1$ close to one (we will take $\gamma_{\rm th} =0.99$ throughout), define the
specialization time
\begin{equation}\label{eq:alpha-spec}
    \alpha_{\rm spec} \;=\; \inf\Big\{\alpha \geq 0 \,:\, \max_{0 \leq s \leq k_s - k_t}
    \gamma_s(\alpha) \geq \gamma_{\rm th}\Big\},
\end{equation}
with $\gamma_s$ given by Eq.~\eqref{eq:gamma}. Then, integrating the ODEs numerically, we find
that, as $\D \to \infty$,
\begin{itemize}
\item[(i)] for $k_s = k_t$, $\alpha_{\rm spec} = \mathcal{O}_{\D}(1)$,
\item[(ii)] for $k_t < k_s \leq P - 2$, $\alpha_{\rm spec} = \mathcal{O}_{\D}(\log \D)$. 
\end{itemize}
Since $\alpha = t/\D$, specialization occurs after $\mathcal{O}_{\D}(\D)$ samples in case (i) and
$\mathcal{O}_{\D}(\D\log\D)$ samples in case (ii).
\end{result}

Result~\ref{res:locality} is obtained for $g(z) = \erf(z/\sqrt 2)$, but we expect it to hold
for any activation with information exponent one \citep{arous2021online}, that is, with a
nonzero first Hermite coefficient, $\E[z\,g(z)] \neq 0$ for $z \sim \mathcal{N}(0, 1)$.
For activations with a higher information exponent, more than $\mathcal{O}_{\D}(\D)$ samples
are expected to be required already at $k_s = k_t$ \citep{arous2021online, abbe2023sgd,
bietti2023learning}, and understanding the role of locality requires a more detailed analysis, which we leave for future work.

\begin{figure}[t]
\centering
\begin{subfigure}{0.48\textwidth}\centering
  \includegraphics[width=\textwidth]{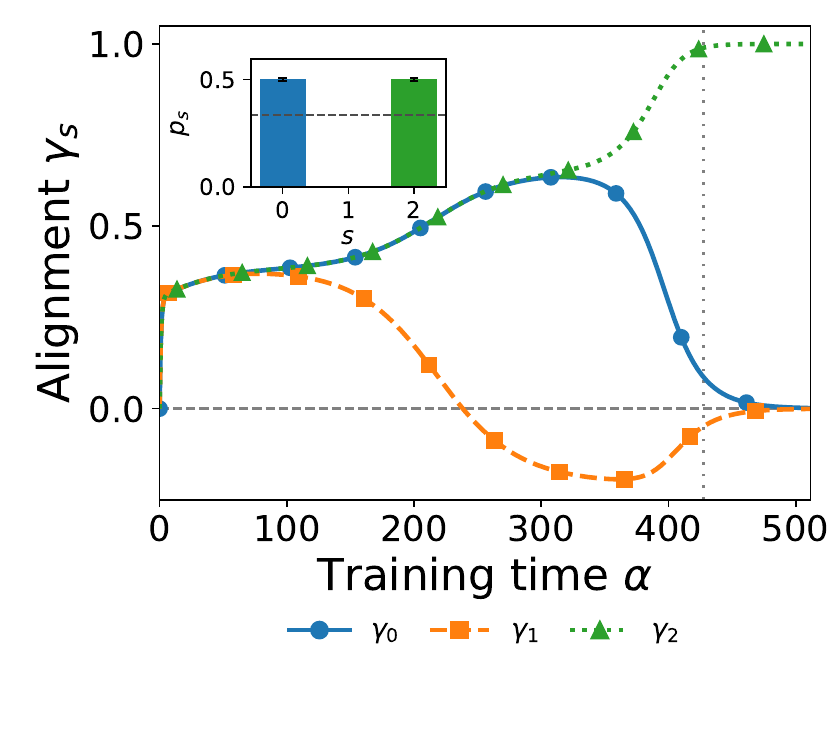}
  \caption{}\label{fig:alignment-a}
\end{subfigure}
\hfill
\begin{subfigure}{0.48\textwidth}\centering
  \includegraphics[width=\textwidth]{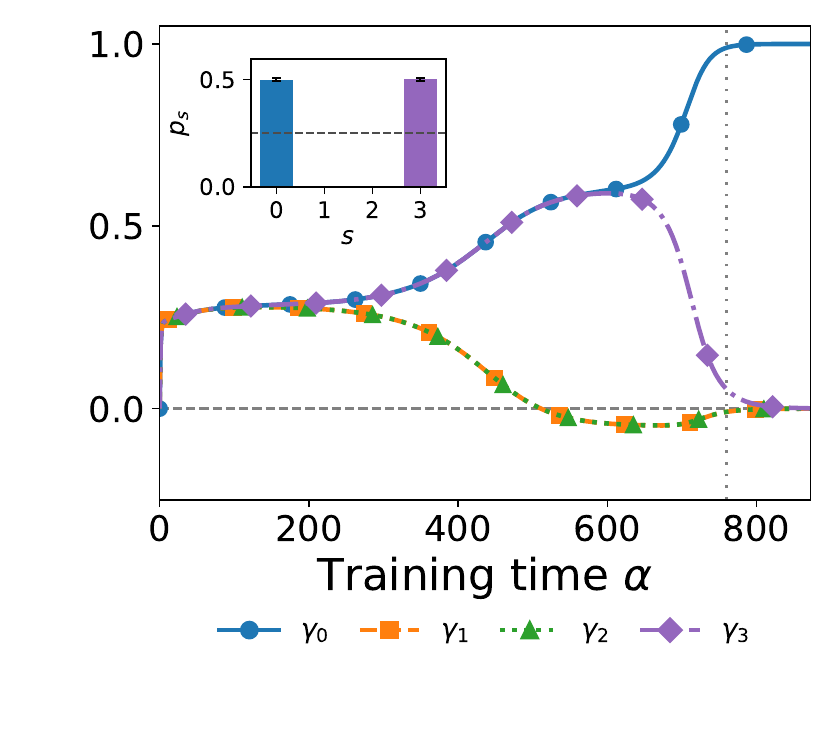}
  \caption{}\label{fig:alignment-b}
\end{subfigure}
\caption{Alignment $\gamma_s$ of the CNN kernel with the solution $s$
vs.~training time $\alpha$, for (a) $k_s = 3$ and (b) $k_s = 4$. The dotted line marks the
specialization time $\alpha_{\rm spec}$ and the dashed line $\gamma_s = 0$. Insets: probability $p_s$ that solution $s$ is selected, over 4096 random
initial conditions (dashed line: uniform). All curves are solutions of the ODEs
\eqref{eq:ODEs-CNN}. Parameters: $\D = 10^{12}$, $g = \erf$, $k_t = 1$, $P = 6$, $S = I$,
$\eta = 0.5$, $T = 1$, $\sigma = 0$.}
\label{fig:alignment}
\end{figure}

Figure~\ref{fig:plateau} shows that for $k_s = k_t$ the error quickly decreases to zero and the specialization time $\alpha_{\rm spec}$ is $D$-independent. For $k_s > k_t$, the generalization error displays a long plateau and then drops abruptly to zero when the student specializes at $\alpha_{\rm spec} = \mathcal{O}_{\D}(\log\D)$. Interestingly, this plateau phase is the result of a particular symmetry of the dynamics. Consider the transformation acting on
the order parameters as
\begin{equation}\label{eq:rho}
    \rho:\quad R_{l l'} \;\mapsto\; R_{k_s - 1 - l,\; k_t - 1 - l'}\,, \qquad
    Q_{l l'} \;\mapsto\; Q_{k_s - 1 - l,\; k_s - 1 - l'}\,.
\end{equation}
In Appendix~\ref{app:reflection} we show that the ODEs \eqref{eq:ODEs-CNN} preserve symmetry under $\rho$. In other words, if the initial condition $(R_0,Q_0)$ is $\rho$-invariant, i.e., $\rho(R_0,Q_0)=(R_0,Q_0)$, then the order parameters remain $\rho$-invariant, $\rho(R(\alpha),Q(\alpha))=(R(\alpha),Q(\alpha))$ for $\alpha>0$. Note that the random initial condition is $\rho$-symmetric up to random fluctuations of order $\D^{-1/2}$, see Eq.~\eqref{eq:init-OP-CNN}.

As shown in Figure~\ref{fig:alignment}, we find that initially the student parameters have approximately equal overlap with different solutions, i.e., $\gamma_s\approx \gamma$ independently of $s$. At intermediate times ($\alpha\sim \mathcal{O}_D(1)$), the student reaches a fixed point with non-uniform overlap with different solutions but that is nevertheless $\rho$-symmetric, so that $\gamma_s=\gamma_{k_s-k_t-s}$. This fixed point corresponds to a plateau with non-zero generalization error. Note that if we were to choose the uninformed initialization $R = 0$, $Q = I/k_s$, which is exactly $\rho$-symmetric, the ODE dynamics would be stuck on this plateau. However, we observe that
the fixed point is unstable
to asymmetric perturbations, which grow as $e^{\kappa\alpha}$ for some $\kappa>0$. Therefore, starting from the initialization asymmetry of order $\D^{-1/2}$, the student escapes the $\rho$-symmetric fixed point after a time of order $\log\D$. 

For $k_s = k_t$ the unique solution is itself $\rho$-symmetric, and the dynamics converges to it without breaking the symmetry, resulting in $\alpha_{\rm spec}\sim 
\mathcal{O}_D(1)$. Note that if $k_s-k_t$ is even, there exists a solution with $s=(k_s-k_t)/2$ that is $\rho$-symmetric. However, we empirically observe that for large enough $D$, this symmetric solution is never reached (see inset of Figure~\ref{fig:alignment-a}).  In Appendix~\ref{app:spec-checks} we show that the plateau and the logarithmic growth of
$\alpha_{\rm spec}$ persist for correlated positions ($c>0$). The specialization transition described here is analogous to that of soft-committee
machines \citep{SaadSolla1995, BiehlSchwarze1995}, where the plateau corresponds to an unspecialized fixed point that is symmetric under permutations of the hidden units.

% =============================================================================
\section{Robustness to spatial correlations}\label{sec:correlations}
% =============================================================================
%\mw{Same here: Early on in the section I would summarize cleanly in one sentence what the impact of weight sharing is.}

So far we have mostly focused on independent positions, $c = 0$. In this section we consider $c > 0$ and investigate how spatial correlations affect the different phases of learning: the beginning of training (Section~\ref{sec:early-time}), the error left after long training (Section~\ref{sec:noise}), and the error at finite training time when the learning rate of each student is optimized (Section~\ref{sec:phase}).
We find that, at fixed learning rate and for $c>0$, the LCN achieves a lower generalization error than the CNN both at early and at late times. However, when the learning rate is appropriately tuned, the LCN wins at early times while the CNN wins at late times, with a transition time that we predict analytically.

\subsection{Early-time expansion}\label{sec:early-time}

In the following result, we characterize the beginning of training by expanding analytically
the ODEs around the random initial conditions in Eqs.~\eqref{eq:init-OP-CNN} and \eqref{eq:init-OP-LCN},
similar to the analysis in \citet{MoriMignacco2025}.

\begin{result}[Early-time expansion]\label{res:early}
Let $S$ be compound symmetric as in Eq.~\eqref{eq:S-compound}, let $k_t \le k_s \le P$, and let
both students start from the random initial conditions in Eqs.~\eqref{eq:init-OP-CNN} and \eqref{eq:init-OP-LCN} at leading order in $\D$.
Then, for small $\alpha$,
\begin{equation}\label{eq:early-expansion}
\begin{aligned}
    \varepsilon_g^{\rm CNN}(\alpha) &= \varepsilon_0^{\rm CNN}
      - \big(\eta A_{\rm CNN} - \eta^2 B_{\rm CNN}\big)\,\alpha + \mathcal{O}(\alpha^2), \\
    \varepsilon_g^{\rm LCN}(\alpha) &= \varepsilon_0^{\rm LCN}
      - \big(\eta A_{\rm LCN} - \eta^2 B_{\rm LCN}\big)\,\alpha + \mathcal{O}(\alpha^2),
\end{aligned}
\end{equation}
with the initial errors
\begin{equation}\label{eq:early-eps0}
\begin{aligned}
    \varepsilon_0^{\rm LCN} &= \frac{2P}{\pi}\left[\arcsin\frac{T}{1 + T} + (P - 1)\arcsin\frac{Tc}{1 + T}\right]
      + \frac{P}{3}, \\
    \varepsilon_0^{\rm CNN} &= \varepsilon_0^{\rm LCN} + \frac{2P(P - 1)}{\pi}\arcsin\frac{c}{2}.
\end{aligned}
\end{equation}
The coefficients $A_{\rm CNN}$, $A_{\rm LCN}$, $B_{\rm CNN}$, and $B_{\rm LCN}$ are positive,
independent of $k_t$, and given in closed form, together with their derivation, in Appendix~\ref{app:early}.
\end{result}

First of all, at initialization the CNN starts from a larger error, $\varepsilon_0^{\rm CNN} > \varepsilon_0^{\rm LCN}$. Indeed, at $c>0$, the preactivations $\hat\lambda_p$ of the CNN are correlated across positions as $\E[\hat\lambda_p\hat\lambda_q] = S_{p,q}$, so that errors in the outputs $g(\hat\lambda_p)$ add up coherently. On the other hand, for the LCN, the parameters at different positions are independent, i.e., $\E[\hat\lambda_p\hat\lambda_q] = \delta_{p,q}$, so the errors in $g(\hat\lambda_p)$ tend to cancel out.

Using Eq.~\eqref{eq:early-expansion}, we can also compare the performance of the two students at small but finite $\alpha$. To remove the effect of the choice of the learning rate, for each student we set the learning rate to the optimal value, $\eta^*_{\rm CNN}=A_{\rm CNN}/(2B_{\rm CNN})$ for the CNN and $\eta^*_{\rm LCN}=A_{\rm LCN}/(2B_{\rm LCN})$ for the LCN, and denote by $\varepsilon^*_{\rm CNN}(\alpha)$ and $\varepsilon^*_{\rm LCN}(\alpha)$ the resulting generalization errors. Then, to leading order in $\alpha$, we find
\begin{equation}
    \begin{cases}
        \varepsilon^*_{\rm CNN} > \varepsilon^*_{\rm LCN} & \text{if } c > c^*(\alpha)\,,\\
        \varepsilon^*_{\rm CNN} < \varepsilon^*_{\rm LCN} & \text{if } c < c^*(\alpha)\,,
    \end{cases}
\end{equation}
where 
\begin{equation}\label{eq:cstar}
    c^*(\alpha) \;=\; c_0\,\alpha + \mathcal{O}(\alpha^2),
\end{equation}
and the slope $c_0$ depends on $P$, $k_s$, $T$, and $\sigma$, and is given in closed form in
Eq.~\eqref{eq:c0} of Appendix~\ref{app:early}. In other words, the CNN achieves lower generalization error for small correlations $c$ (or, equivalently, at long times) while the LCN wins at large $c$ (short times). Although this result is in principle valid only for small $\alpha$, it matches surprisingly well the phase diagram obtained by integrating the exact ODEs numerically (Figure~\ref{fig:phase}).

\subsection{Asymptotic generalization error}\label{sec:noise}

\begin{figure}[t]
    \centering
    \begin{subfigure}{0.32\textwidth}\centering
      \includegraphics[width=\textwidth]{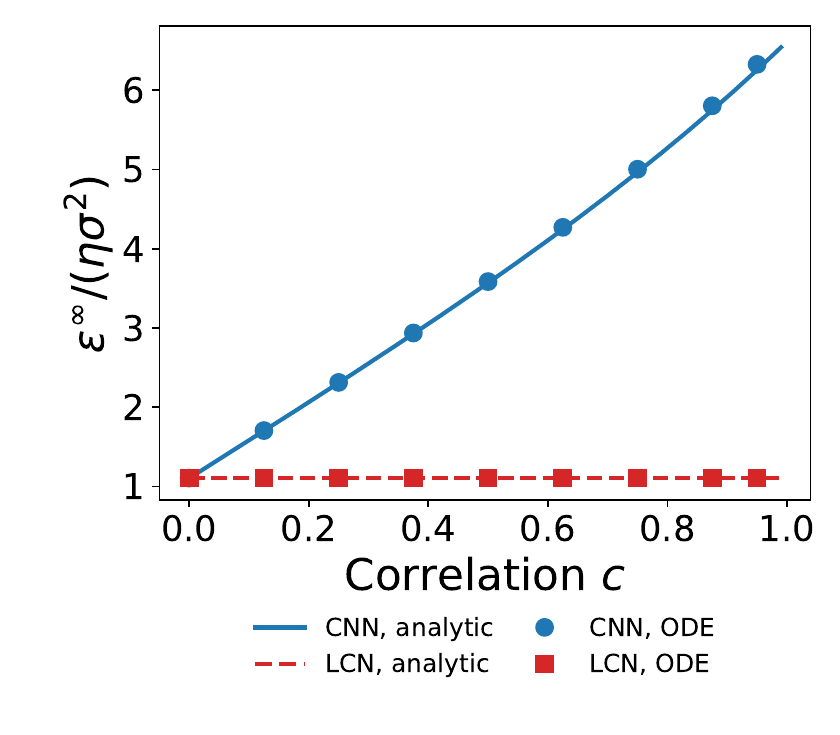}
      \caption{}\label{fig:floor-a}
    \end{subfigure}
    \hfill
    \begin{subfigure}{0.32\textwidth}\centering
      \includegraphics[width=\textwidth]{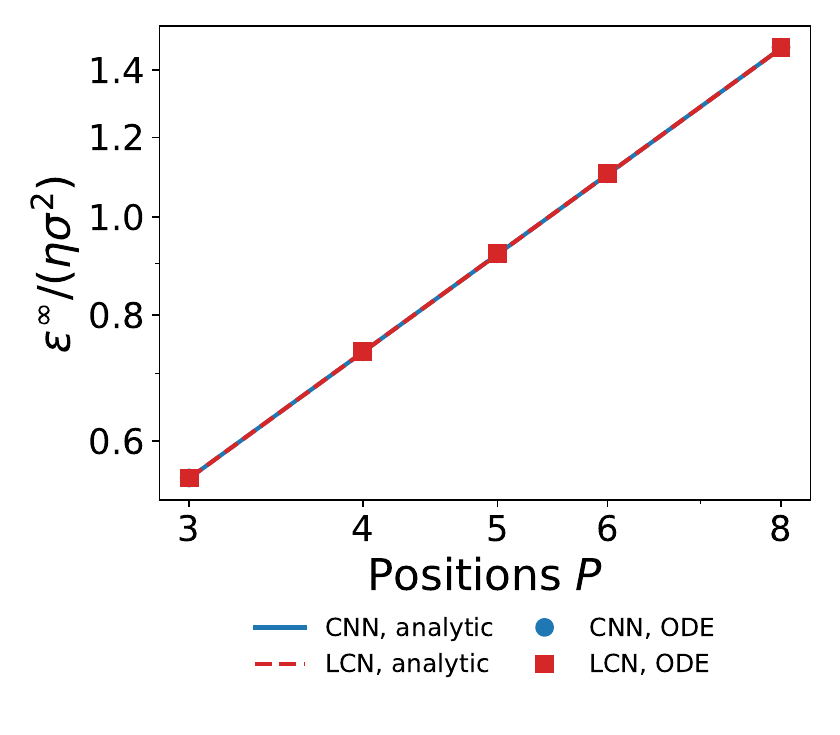}
      \caption{}\label{fig:floor-b}
    \end{subfigure}
    \hfill
    \begin{subfigure}{0.32\textwidth}\centering
      \includegraphics[width=\textwidth]{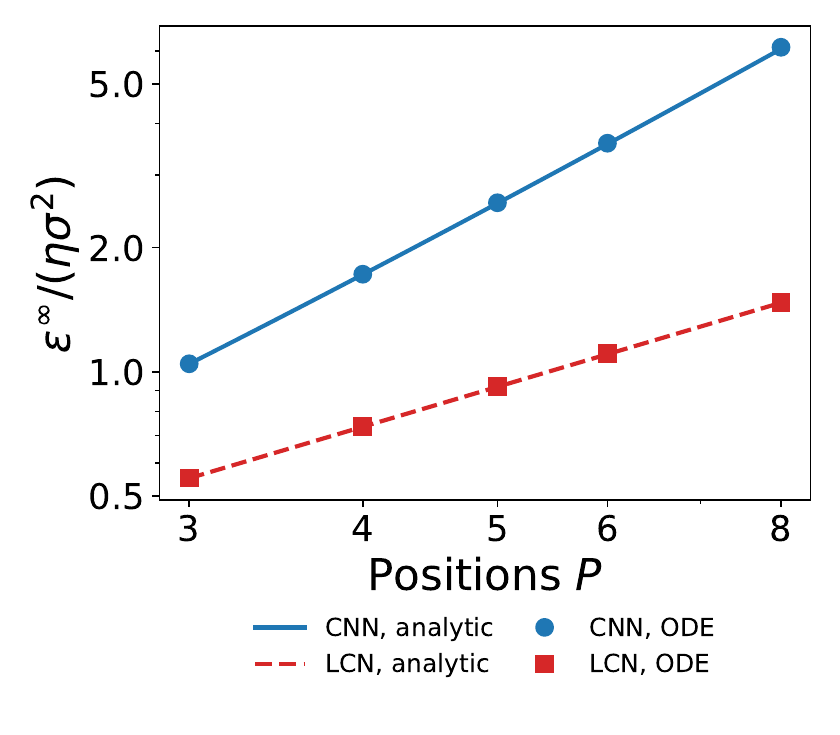}
      \caption{}\label{fig:floor-c}
    \end{subfigure}
    \caption{Asymptotic noise floor $\varepsilon^\infty/(\eta\sigma^2)$ of the CNN
    and the LCN at kernel size $k_s = k_t = 2$, as a function of the data correlation $c$ and the
    number of positions $P$, where $\varepsilon^\infty$ is the asymptotic generalization error
    under label noise. Lines are the closed forms \eqref{eq:eps-inf-A} and
    \eqref{eq:eps-inf-B} and markers are the noise floors located numerically from the
    order-parameter ODEs at finite learning rate. (a) Floor vs.~$c$
    at $P = 6$. The CNN grows steeply with $c$ while the LCN stays flat. (b) Scaling with $P$ at $c =
    0$. The CNN and the LCN coincide and are both linear in $P$. (c) Scaling with $P$ at $c = 0.5$. The LCN
    is still linear in $P$ while the CNN grows quadratically.
    Parameters: $g = \erf$, $k_s = k_t = 2$, $\eta = 2\times 10^{-3}$, $\sigma = 1$, $T = 1$.}
    \label{fig:floor-vs-c}
    \end{figure}

We now turn to long training times and ask how spatial correlations affect the asymptotic error in the presence of label noise, $\sigma > 0$.
We compute this error analytically, through a perturbative expansion of the ODEs at small
learning rate $\eta$.

\begin{result}[Asymptotic generalization error]\label{res:floor}
Let $S$ be compound symmetric with $c \in [0, 1)$, let $\sigma > 0$ and $k_t \le k_s \le P$.
Consider a student that converges to the neighborhood of perfect recovery, that is, of a
zero-error solution of Eq.~\eqref{eq:degenerate}.
To leading order in $\eta$, its asymptotic generalization error, or noise floor,
$\varepsilon^\infty = \lim_{\alpha\to\infty}\varepsilon_g(\alpha)$ is
\begin{equation}\label{eq:eps-inf-A}
    \varepsilon^{\infty,\mathrm{CNN}}(c) \;=\; \frac{\eta\sigma^2}{\pi}\left[\frac{P}{\sqrt{1 + 2T}} + \frac{P(P - 1)\,c}{\sqrt{(1 + T)^2 - T^2 c^2}}\right]+\mathcal{O}(\eta^2),
\end{equation}
\begin{equation}\label{eq:eps-inf-B}
    \varepsilon^{\infty,\mathrm{LCN}} \;=\; \frac{\eta\sigma^2 P}{\pi\sqrt{1 + 2T}}+\mathcal{O}(\eta^2) \,.
\end{equation}
\end{result}

The derivation, given in Appendix~\ref{app:perturbative}, linearizes the ODEs around perfect
recovery and locates the fixed point reached at small $\eta$.
Both expressions are independent of the kernel sizes $k_s$ and $k_t$.
Figure~\ref{fig:floor-vs-c} compares them with the asymptotic errors located numerically from
the ODEs at finite learning rate, finding excellent agreement.
At $c = 0$ the two errors coincide, as expected from Result~\ref{res:rescaling}.
For $c > 0$, the error of the CNN grows with $c$ and quadratically with $P$, while that of the
LCN is independent of $c$ and linear in $P$.
Therefore, at a fixed small learning rate, the LCN is more robust to label noise than the CNN
in the presence of spatial correlations.

To understand the origin of this difference, we consider, without loss of generality, a student
close to the zero-error solution with $s = 0$ in Eq.~\eqref{eq:degenerate}, and define the error
in the preactivation at position $p$,
\begin{equation}\label{eq:e-def}
    e_p \;\equiv\; \hat\lambda_p - \lambda_p ,
\end{equation}
which vanishes at this solution. Near the fixed point we find (Appendix~\ref{app:perturbative})
\begin{equation}\label{eq:err-fields}
    \E\big[e_p^{\mathrm{CNN}} e_q^{\mathrm{CNN}}\big] = \frac{\eta\sigma^2}{2}\,S_{p, q} +\mathcal{O}(\eta^2),
    \qquad
    \E\big[e_p^{\mathrm{LCN}} e_q^{\mathrm{LCN}}\big] = \frac{\eta\sigma^2}{2}\,\delta_{p, q}+\mathcal{O}(\eta^2) \,.
\end{equation}
The errors of the LCN are uncorrelated across positions, because each position has its own
kernel. The CNN instead uses the same kernel at every position, so its errors inherit the
correlations of the data. 
As a result, for $c>0$, the leading-order noise floor of the LCN is lower than that of the CNN when evaluated at the same learning rate. Nevertheless, as we show in Section~\ref{sec:phase}, the CNN achieves a lower generalization error at long training horizons when the learning rate is optimized separately for each architecture.

\subsection{Phase diagram at the optimal learning rate}\label{sec:phase}

\begin{figure}[t]
    \centering
    \includegraphics[width=0.62\textwidth]{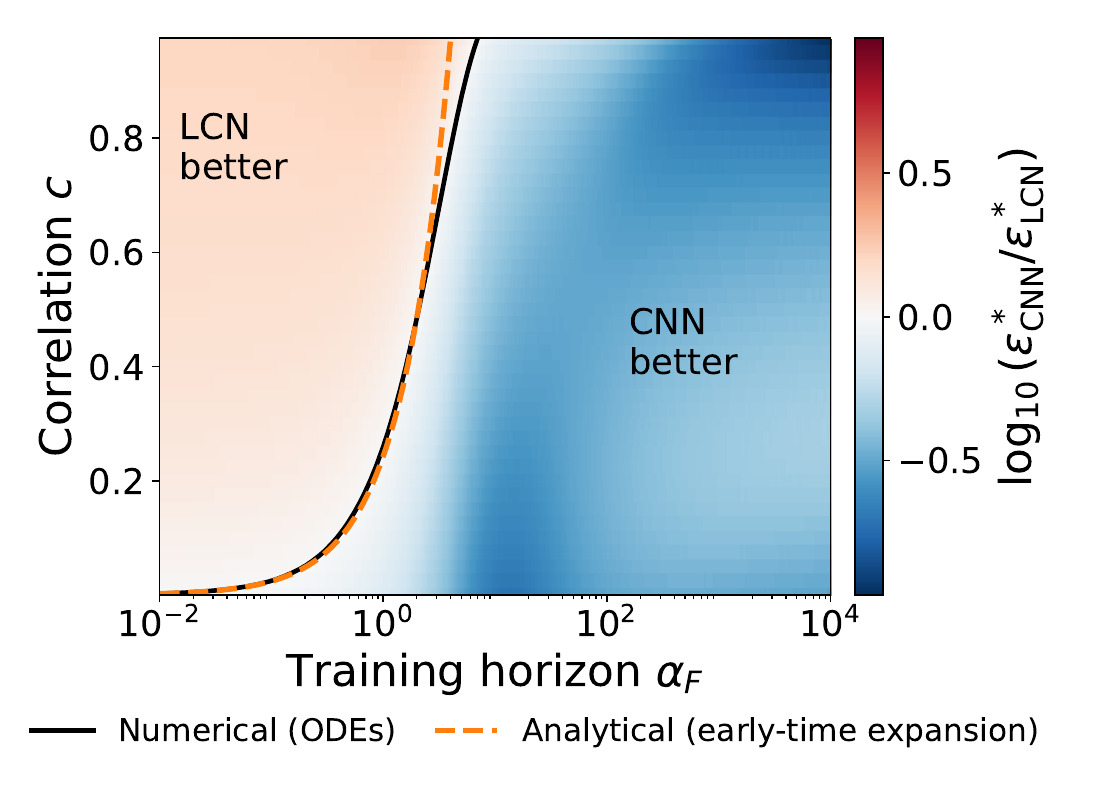}
    \caption{Comparison of the CNN and the LCN at training horizon $\alpha_F$ and correlation $c$, each at its
    optimal constant learning rate. The color shows the log-ratio of the errors
    $\varepsilon^*$.
    Solid line: boundary $\varepsilon^*_{\rm CNN} = \varepsilon^*_{\rm LCN}$, obtained by integrating the
    ODEs numerically.
    Dashed line: analytical prediction Eq.~\eqref{eq:cstar} of the early-time expansion.
    Parameters: $g = \erf$, $P = 4$, $k_s = k_t = 1$, $T = 1$, $\sigma = 1$, random initial
    condition in the limit $\D \to \infty$.}
    \label{fig:phase}
\end{figure}

While Result~\ref{res:floor} compares the asymptotic performance of the two students at the same learning rate, it does not account for two effects. First, the two students will in general converge to perfect recovery at different rates. Moreover, their optimal learning rates can differ and depend on the training horizon.
To include both effects, we use the ODEs~\eqref{eq:ODEs-CNN} and~\eqref{eq:ODEs-LCN} to
compare the two students at a given correlation $c$ and training horizon $\alpha_F$, each with the
constant learning rate $\eta^*$ that minimizes its error at $\alpha_F$ (see Appendix~\ref{app:phase}
for details).

Figure~\ref{fig:phase} shows the resulting phase diagram, obtained by numerically integrating the
ODEs.
The LCN has lower error compared to the CNN for short training horizons and
strong correlations. On the other hand, the CNN wins at long training horizons. We find the same transition for other
values of $P$, $k_s = k_t$, $T$, and $\sigma$ (Figure~\ref{fig:phase-var} in Appendix~\ref{app:phase}). Interestingly, the analytical prediction~\eqref{eq:cstar}, obtained by an early-time expansion,
matches well the numerical boundary between the two regimes.  Note that this does not contradict
Section~\ref{sec:noise}, where the LCN has the lower asymptotic error, because the two
comparisons are made at different learning rates.
Indeed, at a given $\alpha_F$, we find numerically that the optimal learning rate of the CNN is lower than that of the LCN for
every $c > 0$ (Figure~\ref{fig:phase-app-a} in Appendix~\ref{app:phase}). Since the asymptotic error is proportional to
$\eta$ (Result~\ref{res:floor}), the lower learning rate of the CNN compensates for
its larger error at equal $\eta$ (Figure~\ref{fig:phase-app-b}).

\section{Conclusions}\label{sec:conclusions}

In this work, we have considered a teacher--student model to study the online SGD dynamics of
convolutional and locally connected networks. In the limit of large feature dimension, we have
shown that, for uncorrelated positions, weight sharing speeds up learning by exactly a factor
equal to the number of positions. We have also investigated the role of locality, showing numerically that
convolutional students that match the kernel size of the teacher recover it from
$\mathcal{O}_{\D}(\D)$ samples, whereas students with a larger kernel require
$\mathcal{O}_{\D}(\D\log\D)$ samples. Finally, we have investigated the effects of spatial
correlations. We have shown analytically that, at small constant learning rate, the CNN is less robust to
noise in the presence of correlations, and numerically that it outperforms the LCN at long
training horizons when the learning rate is optimized.

Our analysis relies on several simplifying assumptions, which point to natural extensions.
For instance, our inputs are Gaussian and carry no localized structure, and one could
consider data in which the signal is planted in one of the
patches \citep{LahotiKarpWinstonSinghLi2024}. It would also be interesting to extend our
results from regression to classification problems. Moreover, we have considered one-pass
SGD, in which each sample is used only once. Training for multiple passes on a finite
dataset, where overfitting and memorization become relevant, would require other tools, such
as dynamical mean-field theory \citep{mignacco2020dynamical, bordelon2022self,
montanari2025dynamical}.

Several interesting directions remain open. Our equations are valid for an arbitrary covariance $S$ across
positions, and could be used to study correlations that decay with the distance between
positions, as in natural images. Moreover, although we restrict ourselves here to translations,
the theoretical analysis described above could be applied to other groups acting on the
positions. For instance, permutation invariance can be studied with the teacher
\begin{equation}
f^{\star}(x)=\sum_{p=0}^{P-1} g\big(\lambda_p(x)\big), \qquad
\lambda_p(x) = \frac{1}{\sqrt{D}}\Big( w_{\rm s}\cdot x_p
  + w_{\rm o}\cdot \sum_{q\neq p} x_q \Big),
\end{equation}
which is invariant under any permutation of the positions. Finally, we have considered constant
learning rates. Learning-rate schedules could further reduce the error and could be designed with
optimal control \citep{mignacco2026statistical}.

\section*{Acknowledgments}

FM was supported by the Center of Mathematical Sciences and Applications (CMSA) at Harvard
University. MW was partially supported by NSF Award DMS-2406905, an Alfred P. Sloan Fellowship in Mathematics and the AI2050 program at Schmidt Sciences (Grant G-25-69786).

\section*{Statement on the use of AI}

We used the generative AI model Claude Opus 5.5 (Anthropic) to assist with literature review,
deriving and checking analytical expressions, writing the simulation code, and drafting and
revising the text. All results were verified by
the authors, who take full responsibility for the content of this work.

\section*{Code availability}

Code to reproduce the figures is available at
\url{https://github.com/francescomori/CNN_theory}.

\bibliographystyle{plainnat}
\bibliography{paper_draft}

\newpage
\appendix

% =============================================================================
\section{ODE derivation: the CNN}\label{app:ode-A}
% =============================================================================

We derive the deterministic ODE for the order parameters $R$ and $Q$ of the CNN in the
$\D \to \infty$ limit. The derivation is generic in the data covariance $S$ and the kernel
size $k_s$.

\subsection{Setup and building blocks}\label{app:ode-A-setup}

The student weights are updated by online SGD:
\begin{equation}
    \Delta \hat w_l \;=\; -\eta\,\partial_{\hat w_l}\,\ell(x;\theta) \;=\; \frac{\eta\,\delta}{\sqrt{k_s\D}}\sum_{p = 0}^{P - 1} g'(\hat\lambda_p)\,x_{\modP{p + l}}, \qquad \delta = y - \hat y_{\mathrm{CNN}},
\end{equation}
where the sum over $p$ comes from the dependence of $\hat y_{\mathrm{CNN}}$ on $\hat w_l$ through all
$P$ position outputs $g(\hat\lambda_p)$.

For simplicity, in this appendix and in Appendices~\ref{app:ode-B} and~\ref{app:perturbative} we
pad the teacher kernel with zero vectors, $w_l = 0$ for $k_t \le l < k_s$. Then $R$, a
$k_s \times k_t$ matrix in Eq.~\eqref{eq:OP_CNN}, and the teacher overlap matrix $T_{l, l'}$ defined
below become $k_s \times k_s$ matrices whose added entries vanish, and every kernel sum runs over
$\{0, \ldots, k_s - 1\}$.

\paragraph{Single-vector fields.} It is convenient to introduce projections of the input at one position
onto one weight vector,
\begin{equation}
    \mu_r^{(l)} \;\equiv\; \frac{w_l \cdot x_r}{\sqrt{k_t\D}}, \qquad
    \nu_r^{(l)} \;\equiv\; \frac{\hat w_l \cdot x_r}{\sqrt{k_s\D}},
\end{equation}
with which the preactivations decompose simply: $\lambda_p = \sum_{l = 0}^{k_s - 1}
\mu_{\modP{p + l}}^{(l)}$, $\hat\lambda_p = \sum_{l = 0}^{k_s - 1} \nu_{\modP{p +
l}}^{(l)}$, every kernel position sum here and below running over $\{0, \ldots, k_s - 1\}$. The
two-point functions of single-vector fields read
\begin{equation}
    \langle\mu_r^{(l)}\mu_q^{(l')}\rangle = S_{r, q}\,T_{l, l'}, \qquad
    \langle\nu_r^{(l)}\nu_q^{(l')}\rangle = S_{r, q}\,Q_{l, l'}, \qquad
    \langle\mu_r^{(l)}\nu_q^{(l')}\rangle = S_{r, q}\,R_{l', l},
\end{equation}
where the index ordering in the cross term reflects the convention $R_{l, l'} = \hat w_l
\cdot w_{l'} / (\sqrt{k_sk_t}\,\D)$, and $T_{l, l'} = w_l \cdot w_{l'} / (k_t\D)$ is the
teacher's overlap matrix, equal to $(T/k_t)\,\delta_{l, l'}$ for $l, l' < k_t$
by~\eqref{eq:teacher-ortho} and zero otherwise. Expanding one of the two factors with $\lambda_p = \sum_{l = 0}^{k_s - 1}
\mu^{(l)}_{\modP{p + l}}$ and $\hat\lambda_p = \sum_{l = 0}^{k_s - 1} \nu^{(l)}_{\modP{p + l}}$ also gives
the covariances between a preactivation and a single-vector field,
\begin{align}
    \langle\hat\lambda_p\,\mu_r^{(l')}\rangle &= \sum_{l = 0}^{k_s - 1} R_{l, l'}\,S_{\modP{p + l},\,r}, &
    \langle\hat\lambda_p\,\nu_r^{(l')}\rangle &= \sum_{l = 0}^{k_s - 1} Q_{l, l'}\,S_{\modP{p + l},\,r}, \\
    \langle\lambda_q\,\mu_r^{(l')}\rangle &= \sum_{l = 0}^{k_s - 1} T_{l, l'}\,S_{\modP{q + l},\,r}, &
    \langle\lambda_q\,\nu_r^{(l')}\rangle &= \sum_{l = 0}^{k_s - 1} R_{l', l}\,S_{\modP{q + l},\,r}.
\end{align}

\paragraph{Effective covariances.} The preactivations are linear in $x$, so their
covariances follow from the definitions. Inserting \eqref{eq:teacher-pre-act} and using
$\langle x_{p, a}\, x_{q, b}\rangle = S_{p, q}\,\delta_{a, b}$,
\begin{equation}
    \langle\lambda_p \lambda_q\rangle
    \;=\; \frac{1}{k_t\D}\sum_{l, l' = 0}^{k_s - 1} (w_l \cdot w_{l'})\,S_{\modP{p + l},\,\modP{q + l'}}
    \;=\; \sum_{l, l' = 0}^{k_s - 1} T_{l, l'}\,S_{\modP{p + l},\,\modP{q + l'}}.
\end{equation}
The same computation with one or both teacher weight vectors replaced by student ones
gives the remaining covariances, with $T$ replaced by $R$ or $Q$. For any $k_s \times k_s$ matrix
$M$ it is useful to define the effective matrix
\begin{equation}
    M^{\Eff}_{p, q} \;\equiv\; \sum_{l, l' = 0}^{k_s - 1} M_{l, l'}\,S_{\modP{p + l},\,\modP{q + l'}},
\label{eq:Meff}
\end{equation}
a $P \times P$ matrix indexed by positions, so that
\begin{equation}\label{eq:covariances_eff}
    \langle\lambda_p\lambda_q\rangle = T^{\Eff}_{p, q}, \qquad
    \langle\hat\lambda_p\hat\lambda_q\rangle = Q^{\Eff}_{p, q}, \qquad
    \langle\hat\lambda_p\lambda_q\rangle = R^{\Eff}_{p, q}.
\end{equation}

\subsection{Saad--Solla integrals}

The ODEs derived below involve the activation only through a small set of
averages of products of $g$ and $g'$ over jointly Gaussian arguments. These averages, and
the equations built from them, are defined for any activation. What is specific to $g =
\erf$ is that they can be evaluated in closed form, which is why we adopt that choice for
the analytic results. For a $2 \times 2$ covariance $C^{(2)} = (c_{p, q})$,
\begin{equation}
    I_2(C^{(2)}) \;=\; \E[g(u_1) g(u_2)] \;=\; \frac{2}{\pi}\,\arcsin\frac{c_{12}}{\sqrt{(1 + c_{11})(1 + c_{22})}},
\end{equation}
where $C^{(2)}$ is the covariance matrix of $(u_1,u_2)$ and similarly
\begin{align}
    J_2(C^{(2)}) &\;=\; \E[g'(u_1) g'(u_2)] \;=\; \frac{2}{\pi}\,\big[(1 + c_{11})(1 + c_{22}) - c_{12}^2\big]^{-1/2}, \label{eq:J2def}\\
    I_3(C^{(3)}) &\;=\; \E[g'(u_1)\,u_2\,g(u_3)] \;=\; \frac{2}{\pi}\,\frac{c_{23}(1 + c_{11}) - c_{12}c_{13}}{(1 + c_{11})\sqrt{(1 + c_{11})(1 + c_{33}) - c_{13}^2}},
\end{align}
with the slot convention $(u_1, u_2, u_3) = (g', \text{linear}, g)$ for $I_3$. The four-point
integral $I_4$ takes the form
\begin{equation}
    I_4(C^{(4)}) = \E[g'(u_1) g'(u_2) g(u_3) g(u_4)] = \frac{4}{\pi^2}\,\frac{1}{\sqrt{\Lambda_4}}\,\arcsin\frac{\Lambda_0}{\sqrt{\Lambda_1\Lambda_2}},
\end{equation}
where
\begin{align}
    \Lambda_4 &= (1 + c_{11})(1 + c_{22}) - c_{12}^2, \\
    \Lambda_1 &= \Lambda_4(1 + c_{33}) - c_{23}^2(1 + c_{11}) - c_{13}^2(1 + c_{22}) + 2c_{12}c_{13}c_{23}, \\
    \Lambda_2 &= \Lambda_4(1 + c_{44}) - c_{24}^2(1 + c_{11}) - c_{14}^2(1 + c_{22}) + 2c_{12}c_{14}c_{24}, \\
    \Lambda_0 &= \Lambda_4\,c_{34} - c_{23}c_{24}(1 + c_{11}) - c_{13}c_{14}(1 + c_{22}) + c_{12}c_{13}c_{24} + c_{12}c_{14}c_{23}.
\end{align}
Below we write these integrals with their field arguments rather than with the covariance
matrix, as in $J_2(\hat\lambda_p, \hat\lambda_q)$ or $I_3(\hat\lambda_p, \mu_r^{(l')},
\lambda_q)$, understanding that the integral is evaluated at the covariance matrix of the
listed fields. Its entries are the two-point functions of Appendix~\ref{app:ode-A-setup}.

\subsection{ODE for $R$}

We compute $\Delta R_{l, l'} = (\Delta \hat w_l \cdot w_{l'}) / (\sqrt{k_sk_t}\,\D)$ from the SGD update:
\begin{equation}
    \Delta R_{l, l'} \;=\; \frac{\eta\,\delta}{k_s\sqrt{k_t}\,\D\sqrt \D}\sum_{p = 0}^{P - 1} g'(\hat\lambda_p)\,(x_{\modP{p + l}} \cdot w_{l'}) \;=\; \frac{\eta\,\delta}{k_s\D}\sum_{p = 0}^{P - 1} g'(\hat\lambda_p)\,\mu_{\modP{p + l}}^{(l')}.
\end{equation}
Taking the continuum limit with $\alpha = t/\D$, we find
\begin{equation}
    \dot R_{l, l'} \;=\; \frac{\eta}{k_s}\,\sum_{p = 0}^{P - 1} \big\langle\delta\,g'(\hat\lambda_p)\,\mu_{\modP{p + l}}^{(l')}\big\rangle.
\end{equation}
Expanding $\delta = \sum_{q = 0}^{P - 1} g(\lambda_q) + \sigma\xi - \sum_{r = 0}^{P - 1} g(\hat\lambda_r)$ and using
$\langle\xi\rangle = 0$:
\begin{equation}\label{eq:R_A_final}
    \dot R_{l, l'} \;=\; \frac{\eta}{k_s}\,\sum_{p, q = 0}^{P - 1} I_3\!\big(\hat\lambda_p, \mu_{\modP{p + l}}^{(l')}, \lambda_q\big)
    \;-\; \frac{\eta}{k_s}\,\sum_{p, q = 0}^{P - 1} I_3\!\big(\hat\lambda_p, \mu_{\modP{p + l}}^{(l')}, \hat\lambda_q\big).
\end{equation}
\subsection{ODE for $Q$}

Since $Q_{l, l'} = \hat w_l \cdot \hat w_{l'} / (k_s\D)$ involves two student vectors, both are
updated at each step and the increment has three pieces,
\begin{equation}
    \Delta Q_{l, l'} \;=\; \frac{\Delta\hat w_l \cdot \hat w_{l'}}{k_s\D} \;+\; \frac{\hat w_l \cdot \Delta\hat w_{l'}}{k_s\D} \;+\; \frac{\Delta\hat w_l \cdot \Delta\hat w_{l'}}{k_s\D},
\end{equation}
two cross terms of order $\eta$ and one of order $\eta^2$. 

Each cross term repeats the calculation of the previous
subsection with the teacher vector $w_{l'}$ replaced by the student vector $\hat w_{l'}$,
which replaces the single-vector field $\mu^{(l')}$ by $\nu^{(l')}$,
\begin{equation}
    \frac{\Delta\hat w_l \cdot \hat w_{l'}}{k_s\D} = \frac{\eta\,\delta}{k_s\D}\sum_{p = 0}^{P - 1} g'(\hat\lambda_p)\,\nu_{\modP{p + l}}^{(l')},
    \qquad
    \frac{\hat w_l \cdot \Delta\hat w_{l'}}{k_s\D} = \frac{\eta\,\delta}{k_s\D}\sum_{p = 0}^{P - 1} g'(\hat\lambda_p)\,\nu_{\modP{p + l'}}^{(l)}.
\end{equation}
Expanding $\delta$ as before and averaging, the two terms give
\begin{align}
    \dot Q_{l, l'}\Big|_{\mathcal{O}(\eta)} \;=\;& \frac{\eta}{k_s}\sum_{p, q = 0}^{P - 1}\Big[I_3\big(\hat\lambda_p, \nu_{\modP{p + l}}^{(l')}, \lambda_q\big) - I_3\big(\hat\lambda_p, \nu_{\modP{p + l}}^{(l')}, \hat\lambda_q\big)\Big] \nonumber \\
    +\;& \frac{\eta}{k_s}\sum_{p, q = 0}^{P - 1}\Big[I_3\big(\hat\lambda_p, \nu_{\modP{p + l'}}^{(l)}, \lambda_q\big) - I_3\big(\hat\lambda_p, \nu_{\modP{p + l'}}^{(l)}, \hat\lambda_q\big)\Big].
\end{align}
The remaining piece is
\begin{equation}
    \frac{\Delta\hat w_l \cdot \Delta\hat w_{l'}}{k_s\D} = \frac{\eta^2\delta^2}{k_s^2\D^2}\sum_{p, q = 0}^{P - 1} g'(\hat\lambda_p)\,g'(\hat\lambda_q)\,\big(x_{\modP{p + l}} \cdot x_{\modP{q + l'}}\big).
\end{equation}
The inner product $x_r \cdot x_q$ is a sum of $\D$ terms, hence of order $\D$, and it
concentrates: its mean is $\D\,S_{r, q}$ and its fluctuations are of order $\sqrt\D$. At
leading order in $\D$ it may therefore be replaced by its average and factored out of the
remaining average,
\begin{equation}
    \big\langle (x_r \cdot x_q)\, f(x) \big\rangle \;=\; \D\,S_{r, q}\,\langle f(x)\rangle \;+\; \mathcal{O}_{\D}(\sqrt\D).
\end{equation}
Applying this with $f(x) = \delta^2 g'(\hat\lambda_p) g'(\hat\lambda_q)$, we find
\begin{equation}
    \big\langle\Delta Q_{l, l'}\big\rangle\Big|_{\mathcal{O}(\eta^2)} = \frac{\eta^2}{k_s^2\D}\sum_{p, q = 0}^{P - 1} S_{\modP{p + l},\,\modP{q + l'}}\,\big\langle\delta^2\,g'(\hat\lambda_p)g'(\hat\lambda_q)\big\rangle.
\end{equation}
Splitting the residual $\delta$ into its noiseless part and the
label noise, $\delta = \Delta y + \sigma\xi$ with $\Delta y = \sum_{r = 0}^{P - 1} g(\lambda_r) - \sum_{r = 0}^{P - 1}
g(\hat\lambda_r)$, and using that $\xi$ is centered and independent of $x$, the cross term
drops and
\begin{equation}
    \big\langle\delta^2 g'(\hat\lambda_p)g'(\hat\lambda_q)\big\rangle = \big\langle(\Delta y)^2 g'(\hat\lambda_p)g'(\hat\lambda_q)\big\rangle + \sigma^2 J_2\big(\hat\lambda_p, \hat\lambda_q\big).
\end{equation}
Expanding the square, we find
\begin{equation}
    (\Delta y)^2 = \sum_{r, s = 0}^{P - 1}\Big[g(\lambda_r)g(\lambda_s) - 2\,g(\lambda_r)g(\hat\lambda_s) + g(\hat\lambda_r)g(\hat\lambda_s)\Big],
\end{equation}
and therefore
\begin{align}
    \big\langle(\Delta y)^2 g'(\hat\lambda_p)g'(\hat\lambda_q)\big\rangle = \sum_{r, s = 0}^{P - 1}\Big[&I_4\big(\hat\lambda_p, \hat\lambda_q, \lambda_r, \lambda_s\big) - 2\,I_4\big(\hat\lambda_p, \hat\lambda_q, \lambda_r, \hat\lambda_s\big) \nonumber \\
    &+ I_4\big(\hat\lambda_p, \hat\lambda_q, \hat\lambda_r, \hat\lambda_s\big)\Big],
\end{align}
with the slot convention $(g', g', g, g)$. Every entry of the $3 \times 3$ and $4 \times
4$ covariance matrices is read off from the two-point functions of Appendix~\ref{app:ode-A-setup}.

Finally, the equation for $Q$ reads 
\begin{align}\label{eq:dQ-full}
    \dot Q_{l, l'} \;=\;& \frac{\eta}{k_s} \sum_{p, q = 0}^{P - 1} \Big[I_3\big(\hat\lambda_p, \nu_{\modP{p + l}}^{(l')}, \lambda_q\big) - I_3\big(\hat\lambda_p, \nu_{\modP{p + l}}^{(l')}, \hat\lambda_q\big)\Big] \nonumber \\
    +\;& \frac{\eta}{k_s} \sum_{p, q = 0}^{P - 1} \Big[I_3\big(\hat\lambda_p, \nu_{\modP{p + l'}}^{(l)}, \lambda_q\big) - I_3\big(\hat\lambda_p, \nu_{\modP{p + l'}}^{(l)}, \hat\lambda_q\big)\Big] \nonumber \\
    +\;& \frac{\eta^2}{k_s^2} \sum_{p, q = 0}^{P - 1} S_{\modP{p + l},\,\modP{q + l'}}\bigg\{\sum_{r, s = 0}^{P - 1}\Big[I_4\big(\hat\lambda_p, \hat\lambda_q, \lambda_r, \lambda_s\big) \nonumber \\
    &\hspace{4.2em} - 2\,I_4\big(\hat\lambda_p, \hat\lambda_q, \lambda_r, \hat\lambda_s\big) + I_4\big(\hat\lambda_p, \hat\lambda_q, \hat\lambda_r, \hat\lambda_s\big)\Big] + \sigma^2 J_2\big(\hat\lambda_p, \hat\lambda_q\big)\bigg\} ,
\end{align}
with all position sums running over $\{0, \ldots, P - 1\}$.

Together with the equation for $R$, this is the closed system~\eqref{eq:ODEs-CNN}, with
\begin{align}
    \big[F_R^{\rm CNN}\big]_{l, l'} \;=\;& \sum_{p, q = 0}^{P - 1} \Big[I_3\big(\hat\lambda_p, \mu_{\modP{p + l}}^{(l')}, \lambda_q\big) - I_3\big(\hat\lambda_p, \mu_{\modP{p + l}}^{(l')}, \hat\lambda_q\big)\Big], \label{eq:FR-cnn} \\
    \big[F_Q^{\rm CNN}\big]_{l, l'} \;=\;& \sum_{p, q = 0}^{P - 1} \Big[I_3\big(\hat\lambda_p, \nu_{\modP{p + l}}^{(l')}, \lambda_q\big) - I_3\big(\hat\lambda_p, \nu_{\modP{p + l}}^{(l')}, \hat\lambda_q\big)\Big] + (l \leftrightarrow l'), \label{eq:FQ-cnn} \\
    \big[G_Q^{\rm CNN}\big]_{l, l'} \;=\;& \sum_{p, q = 0}^{P - 1} S_{\modP{p + l},\,\modP{q + l'}}\bigg\{\sum_{r, s = 0}^{P - 1}\Big[I_4\big(\hat\lambda_p, \hat\lambda_q, \lambda_r, \lambda_s\big) - 2\,I_4\big(\hat\lambda_p, \hat\lambda_q, \lambda_r, \hat\lambda_s\big) \nonumber \\
    &\hspace{4.2em} + I_4\big(\hat\lambda_p, \hat\lambda_q, \hat\lambda_r, \hat\lambda_s\big)\Big] + \sigma^2 J_2\big(\hat\lambda_p, \hat\lambda_q\big)\bigg\}. \label{eq:GQ-cnn}
\end{align}
These functions depend on $R$, $Q$, $T$, and $S$ through the covariance matrices at which
the integrals are evaluated, given in Appendix~\ref{app:ode-A-setup}.

% =============================================================================
\section{ODE derivation: the LCN}\label{app:ode-B}
% =============================================================================

Throughout this appendix $\hat\lambda$ denotes the LCN's preactivation. The derivation
parallels Appendix~\ref{app:ode-A} but with two key
differences: (i) each per-position kernel $\hat w^{(p)}$ is updated only from gradients at
its own position $p$, removing the position sum, and (ii) the state now includes per-pair
student--student blocks $Q^{(p, q)}$, which couple the dynamics across positions when the
windows of $p$ and $q$ overlap.

\subsection{Setup}

As in Appendix~\ref{app:ode-A}, we pad the teacher kernel with zero vectors, so that each
$R^{(p)}$ is a $k_s \times k_s$ matrix whose added entries vanish.

The SGD update for the $p$-th kernel is
\begin{equation}
    \Delta \hat w^{(p)}_l \;=\; \frac{\eta\,\delta}{\sqrt{k_s\D}}\,g'(\hat\lambda_p)\,x_{\modP{p + l}}.
    \label{eq:update-B}
\end{equation}
The single-vector fields are now indexed by position and by which weight vector of
the kernel is used: $\nu_r^{(p, l)}
\equiv \hat w^{(p)}_l \cdot x_r / \sqrt{k_s\D}$, and $\mu_r^{(l)} \equiv w_l \cdot x_r /
\sqrt{k_t\D}$ as before. The two-point functions are
\begin{equation}
    \langle\mu_r^{(l)}\nu_s^{(p, l')}\rangle = S_{r, s}\,R^{(p)}_{l', l}, \qquad
    \langle\nu_r^{(p, l)}\nu_s^{(q, l')}\rangle = S_{r, s}\,Q^{(p, q)}_{l, l'},
\end{equation}
where, as in Appendix~\ref{app:ode-A}, the indices on $R^{(p)}$ appear in the opposite
order to the slots because $\mu$ carries a teacher index and $\nu$ a student one, while
the convention is $R^{(p)}_{l, l'} = \hat w^{(p)}_l \cdot w_{l'} / (\sqrt{k_sk_t}\,\D)$. Expanding
$\hat\lambda_p = \sum_{l'' = 0}^{k_s - 1} \nu^{(p, l'')}_{\modP{p + l''}}$ and $\lambda_s =
\sum_{l'' = 0}^{k_s - 1} \mu^{(l'')}_{\modP{s + l''}}$ gives the covariances between a local
field and a single-vector field,
\begin{align}
    \langle\hat\lambda_p\,\mu_r^{(l')}\rangle &= \sum_{l'' = 0}^{k_s - 1} R^{(p)}_{l'', l'}\,S_{\modP{p + l''},\,r}, &
    \langle\hat\lambda_p\,\nu_r^{(q, l')}\rangle &= \sum_{l'' = 0}^{k_s - 1} Q^{(p, q)}_{l'', l'}\,S_{\modP{p + l''},\,r}, \\
    \langle\lambda_s\,\mu_r^{(l')}\rangle &= \sum_{l'' = 0}^{k_s - 1} T_{l', l''}\,S_{\modP{s + l''},\,r}, &
    \langle\lambda_s\,\nu_r^{(q, l')}\rangle &= \sum_{l'' = 0}^{k_s - 1} R^{(q)}_{l', l''}\,S_{\modP{s + l''},\,r},
\end{align}
the third of which involves the teacher alone and is unchanged from
Appendix~\ref{app:ode-A}. The preactivation covariances generalize to
\begin{equation}\label{eq:covariances_eff_LCN}
    \langle\hat\lambda_p \hat\lambda_q\rangle = \sum_{l, l' = 0}^{k_s - 1} Q^{(p, q)}_{l, l'}\,S_{\modP{p + l},\,\modP{q + l'}}, \qquad
    \langle\hat\lambda_p \lambda_q\rangle = \sum_{l, l' = 0}^{k_s - 1} R^{(p)}_{l, l'}\,S_{\modP{p + l},\,\modP{q + l'}}.
\end{equation}

\subsection{ODEs}

The ODEs follow by direct computation, in the same way as for the CNN:
\begin{align}
    \dot R^{(p)}_{l, l'} &\;=\; \frac{\eta}{k_s}\Big[\sum_{s = 0}^{P - 1} I_3(\hat\lambda_p, \mu_{\modP{p + l}}^{(l')}, \lambda_s) - \sum_{q = 0}^{P - 1} I_3(\hat\lambda_p, \mu_{\modP{p + l}}^{(l')}, \hat\lambda_q)\Big], \\
    \dot Q^{(p, q)}_{l, l'}\Big|_{\mathcal{O}(\eta)} &\;=\; \frac{\eta}{k_s}\,\Big\{\sum_{s = 0}^{P - 1} I_3(\hat\lambda_p, \nu_{\modP{p + l}}^{(q, l')}, \lambda_s) - \sum_{r = 0}^{P - 1} I_3(\hat\lambda_p, \nu_{\modP{p + l}}^{(q, l')}, \hat\lambda_r)\Big\} + (p \leftrightarrow q,\,l \leftrightarrow l'), \\
    \dot Q^{(p, q)}_{l, l'}\Big|_{\mathcal{O}(\eta^2)} &\;=\; \frac{\eta^2}{k_s^2}\,S_{\modP{p + l},\,\modP{q + l'}}\,\Big[\langle(\Delta y)^2\,g'(\hat\lambda_p)g'(\hat\lambda_q)\rangle + \sigma^2 J_2\big(\hat\lambda_p, \hat\lambda_q\big)\Big].
\end{align}
The structure is identical to the CNN's but for the absence of the position sum in the $\dot
R^{(p)}$ equation (the $p$-th kernel only feels gradients from position $p$). This is the
structural origin of the factor-of-$P$ rescaling between the CNN and the LCN at $S = I$.

Expanding the weighted-error term as in
Appendix~\ref{app:ode-A}, the equation for $Q^{(p, q)}$ reads in full
\begin{align}\label{eq:dQ-full-B}
    \dot Q^{(p, q)}_{l, l'} \;=\;& \frac{\eta}{k_s} \Big[\sum_{s = 0}^{P - 1} I_3\big(\hat\lambda_p, \nu_{\modP{p + l}}^{(q, l')}, \lambda_s\big) - \sum_{r = 0}^{P - 1} I_3\big(\hat\lambda_p, \nu_{\modP{p + l}}^{(q, l')}, \hat\lambda_r\big)\Big] \nonumber \\
    +\;& \frac{\eta}{k_s} \Big[\sum_{s = 0}^{P - 1} I_3\big(\hat\lambda_q, \nu_{\modP{q + l'}}^{(p, l)}, \lambda_s\big) - \sum_{r = 0}^{P - 1} I_3\big(\hat\lambda_q, \nu_{\modP{q + l'}}^{(p, l)}, \hat\lambda_r\big)\Big] \nonumber \\
    +\;& \frac{\eta^2}{k_s^2}\, S_{\modP{p + l},\,\modP{q + l'}}\bigg\{\sum_{r, s = 0}^{P - 1}\Big[I_4\big(\hat\lambda_p, \hat\lambda_q, \lambda_r, \lambda_s\big) \nonumber \\
    &\hspace{5.4em} - 2\,I_4\big(\hat\lambda_p, \hat\lambda_q, \lambda_r, \hat\lambda_s\big) + I_4\big(\hat\lambda_p, \hat\lambda_q, \hat\lambda_r, \hat\lambda_s\big)\Big] + \sigma^2 J_2\big(\hat\lambda_p, \hat\lambda_q\big)\bigg\} ,
\end{align}
with all position sums running over $\{0, \ldots, P - 1\}$. Comparing with
\eqref{eq:dQ-full}, the only structural difference is that the CNN carries a further sum over
positions in each term, and a double sum in the $\mathcal{O}(\eta^2)$ prefactor, where the LCN carries a
single $S$ factor.

In the notation of~\eqref{eq:ODEs-LCN}, the LCN dynamics is therefore given by
\begin{align}
    \big[F_R^{\rm LCN}\big]^{(p)}_{l, l'} \;=\;& \sum_{s = 0}^{P - 1} I_3\big(\hat\lambda_p, \mu_{\modP{p + l}}^{(l')}, \lambda_s\big) - \sum_{q = 0}^{P - 1} I_3\big(\hat\lambda_p, \mu_{\modP{p + l}}^{(l')}, \hat\lambda_q\big), \label{eq:FR-lcn} \\
    \big[F_Q^{\rm LCN}\big]^{(p, q)}_{l, l'} \;=\;& \sum_{s = 0}^{P - 1} I_3\big(\hat\lambda_p, \nu_{\modP{p + l}}^{(q, l')}, \lambda_s\big) - \sum_{r = 0}^{P - 1} I_3\big(\hat\lambda_p, \nu_{\modP{p + l}}^{(q, l')}, \hat\lambda_r\big) \nonumber \\
    &+ (p \leftrightarrow q,\, l \leftrightarrow l'), \label{eq:FQ-lcn} \\
    \big[G_Q^{\rm LCN}\big]^{(p, q)}_{l, l'} \;=\;& S_{\modP{p + l},\,\modP{q + l'}}\bigg\{\sum_{r, s = 0}^{P - 1}\Big[I_4\big(\hat\lambda_p, \hat\lambda_q, \lambda_r, \lambda_s\big) - 2\,I_4\big(\hat\lambda_p, \hat\lambda_q, \lambda_r, \hat\lambda_s\big) \nonumber \\
    &\hspace{4.2em} + I_4\big(\hat\lambda_p, \hat\lambda_q, \hat\lambda_r, \hat\lambda_s\big)\Big] + \sigma^2 J_2\big(\hat\lambda_p, \hat\lambda_q\big)\bigg\}, \label{eq:GQ-lcn}
\end{align}
with the integrals evaluated at the LCN covariances given above.

% =============================================================================
\section{Generalization error in terms of order parameters}\label{app:gen-error}
% =============================================================================

For either of the two students, the population squared error against the clean teacher can be written
\begin{equation}
    \varepsilon_g \;=\; \E\big[\big(f^\star(x) - \hat y(x)\big)^2\big] \;=\; \E[f^{\star 2}] - 2\,\E[f^\star \hat y] + \E[\hat y^2].
\end{equation}
Expanding $f^\star = \sum_{p = 0}^{P - 1} g(\lambda_p)$ and $\hat y = \sum_{r = 0}^{P - 1}
g(\hat\lambda_r)$, each of the three averages becomes a sum of two-point integrals of the
type $\E[g(u_1) g(u_2)]$ with $(u_1, u_2)$ jointly Gaussian, so that
\begin{equation}\label{eq:gen-error-I2}
    \varepsilon_g \;=\; \sum_{p, q = 0}^{P - 1} I_2\big(\lambda_p, \lambda_q\big)
    \;-\; 2\sum_{p, r = 0}^{P - 1} I_2\big(\lambda_p, \hat\lambda_r\big)
    \;+\; \sum_{r, s = 0}^{P - 1} I_2\big(\hat\lambda_r, \hat\lambda_s\big),
\end{equation}
where the integrals are written with their field arguments, as elsewhere in the appendices.
Here $\hat\lambda_r$ stands for $\hat\lambda^{\mathrm{CNN}}_r$ or $\hat\lambda^{\mathrm{LCN}}_r$
according to the student, the index $r$ running over positions in both cases. The expression
is thus the same for the two architectures, which
enter only through the preactivation covariances
$\langle\lambda_p\lambda_q\rangle$, $\langle\lambda_p\hat\lambda_r\rangle$, and
$\langle\hat\lambda_r\hat\lambda_s\rangle$ at which the integrals are evaluated. These are
given in terms of the order parameters in Appendices~\ref{app:ode-A} and~\ref{app:ode-B}.

% =============================================================================
\section{Exact $P$-fold rescaling at $S = I$}\label{app:rescaling}

In this appendix, we show that if $S=I$, $k_t \le k_s\leq P$, and the CNN and the LCN have coinciding initial
conditions, the dynamics of the CNN is exactly a $P$-fold rescaling of
the dynamics of the LCN. Every step below involves only the student indices, so the
argument does not require the student and teacher kernels to have the same size.
We assume that the initial conditions of the LCN are position-symmetric, meaning that 
$R^{(p)}$ is independent of $p$ and $Q^{(p,q)}$ depends on $p$ and $q$ only through $\modP{q-p}$.

As a first step, we show that the generalization error of the LCN only depends on a subset of its order
parameters. Indeed, the generalization error depends on the covariance
\begin{equation}\label{eq:cov}
    \langle\hat\lambda^{\mathrm{LCN}}_p \hat\lambda^{\mathrm{LCN}}_q\rangle
    \;=\; \sum_{l, l' = 0}^{k_s - 1} Q^{(p, q)}_{l l'}\, S_{\modP{p + l},\,\modP{q + l'}}
  =
    \sum_{l, l' = 0}^{k_s - 1} Q^{(p, q)}_{l l'}\, \delta_{\modP{p + l},\,\modP{q + l'}} .
\end{equation}
This leads us to call an entry $(p, q, l, l')$ of $Q$ \emph{aligned} if
\begin{equation}\label{eq:aligned}
    \modP{p + l} \;=\; \modP{q + l'},
    \qquad\text{equivalently}\qquad
    q \;=\; \modP{p + l - l'} .
\end{equation}
Given $(p, l, l')$, the condition \eqref{eq:aligned} determines $q = \modP{p + l - l'}$
uniquely. Note that the aligned component of $Q$ contains $P k_s^2$ elements 
out of the $P^2 k_s^2$ entries of $Q$. On the other hand, the covariance $\langle\hat\lambda^{\mathrm{LCN}}_p \lambda_s\rangle$ 
involves the full matrix $R$. We will next show that the dynamics of the aligned component 
of $Q$ and $R$ can be closed, so that we do not need to track the non-aligned component of $Q$. 
Under position-symmetric initialization, we will show that we can further reduce the 
description of the LCN by a factor $P$: $R$ is described by $k_s k_t$ variables and the aligned part of $Q$ by $k_s^2$ variables. 
Finally, we will show that after rescaling time by a factor $P$, the dynamics of the CNN and the LCN coincide.

To show that the dynamics of $R$ and the aligned part of $Q$ is closed, it is sufficient to
verify that each term of the ODEs does not depend on the non-aligned part of $Q$. Looking at the
evolution equation for $Q^{(p,q)}_{l l'}$, where $q=\modP{p + l - l'}$, there are only two non-trivial terms.
The first one is
\begin{equation}
    \big\langle\hat\lambda_p^{\mathrm{LCN}}\, \nu^{(q, l')}_{\modP{p + l}}\big\rangle
    = \sum_{l'' = 0}^{k_s - 1} Q^{(p, q)}_{l'' l'}\, \delta_{\modP{p + l''},\,\modP{p + l}}
    = Q^{(p, q)}_{l l'} ,
\end{equation}
resulting in the entry itself. The second one is
\begin{equation}\label{eq:key}
    \big\langle \nu^{(q, l')}_{\modP{p + l}}\, \hat\lambda_r^{\mathrm{LCN}} \big\rangle
    = \sum_{l'' = 0}^{k_s - 1} Q^{(q, r)}_{l' l''}\, \delta_{\modP{r + l''},\,\modP{p + l}}
    = Q^{(q, r)}_{l' l''^{\star}}, \qquad l''^{\star} = \modP{p + l - r},
\end{equation}
if $l''^{\star} \in \{0, \dots, k_s-1\}$, and zero otherwise. It is easy to verify that $Q^{(q, r)}_{l' l''^{\star}}$ is aligned.
All terms in the ODE for $R$ are trivially independent of the non-aligned part of $Q$.

When $S=I$ and the LCN is initialized position-symmetrically (for instance, this is valid approximately
if the LCN is initialized at random), it is useful to introduce the reduced position-symmetric ansatz
\begin{align}
    R^{(p)}_{ll'}=\tilde{R}_{ll'}\,, \quad Q^{(p,\modP{p+l-l'})}_{ll'}=\tilde{Q}_{ll'}\,,
\end{align}
consisting of a $k_s\times k_t$ and a $k_s\times k_s$ matrix (as for the order parameters of the CNN). We will next show that 
this ansatz satisfies the ODEs of the LCN and that, after rescaling time by a factor $P$, these equations 
coincide with those of the CNN with the identification $\tilde{R}_{ll'}=R_{ll'}$ and $\tilde{Q}_{ll'}=Q_{ll'}$, 
where $Q$ and $R$ are the order parameters of the CNN.

Substituting $R^{(p)}_{ll'}=\tilde{R}_{ll'}$ and $Q^{(p,\modP{p+l-l'})}_{ll'}=\tilde{Q}_{ll'}$, we get
\begin{align}\label{eq:R_reduced_B}
    \dot{\tilde{R}}_{l, l'} &\;=\; \frac{\eta}{k_s}\sum_{q = 0}^{P - 1} \Big[ I_3(\hat\lambda_p^{\mathrm{LCN}}, \mu_{\modP{p + l}}^{(l')}, \lambda_q) -  I_3(\hat\lambda_p^{\mathrm{LCN}}, \mu_{\modP{p + l}}^{(l')}, \hat\lambda_q^{\mathrm{LCN}})\Big], 
\end{align}
where now $p$ can be any fixed position index. It is easy to check that the right-hand side is $p$-independent
under the position-symmetric ansatz. Similarly, 
\begin{align}\label{eq:Q_reduced_B}
    \dot{\tilde{Q}}_{l, l'} \;=\;& \frac{\eta}{k_s}\sum_{r = 0}^{P - 1} \Big[ I_3\big(\hat\lambda_p^{\mathrm{LCN}}, \nu_{\modP{p + l}}^{(q, l')}, \lambda_r\big) - I_3\big(\hat\lambda_p^{\mathrm{LCN}}, \nu_{\modP{p + l}}^{(q, l')}, \hat\lambda_r^{\mathrm{LCN}}\big)\Big] \nonumber \\
    +\;& \frac{\eta}{k_s} \sum_{r = 0}^{P - 1} \Big[ I_3\big(\hat\lambda_q^{\mathrm{LCN}}, \nu_{\modP{q + l'}}^{(p, l)}, \lambda_r\big) -  I_3\big(\hat\lambda_q^{\mathrm{LCN}}, \nu_{\modP{q + l'}}^{(p, l)}, \hat\lambda_r^{\mathrm{LCN}}\big)\Big] \nonumber \\
    +\;& \frac{\eta^2}{k_s^2}\, \bigg\{\sum_{r, s = 0}^{P - 1}\Big[I_4\big(\hat\lambda_p^{\mathrm{LCN}}, \hat\lambda_q^{\mathrm{LCN}}, \lambda_r, \lambda_s\big)- 2\,I_4\big(\hat\lambda_p^{\mathrm{LCN}}, \hat\lambda_q^{\mathrm{LCN}}, \lambda_r, \hat\lambda_s^{\mathrm{LCN}}\big)  \nonumber \\
    &\hspace{5.4em} + I_4\big(\hat\lambda_p^{\mathrm{LCN}}, \hat\lambda_q^{\mathrm{LCN}}, \hat\lambda_r^{\mathrm{LCN}}, \hat\lambda_s^{\mathrm{LCN}}\big)\Big] + \sigma^2 J_2\big(\hat\lambda_p^{\mathrm{LCN}}, \hat\lambda_q^{\mathrm{LCN}}\big)\bigg\} ,
\end{align}
where again $p$ is any fixed position index and $q=\modP{p+l-l'}$. The right-hand side again does not depend on $p$.

Finally, assuming $S=I$, the ODEs for the CNN can also be simplified:
\begin{align}\label{eq:R_reduced_A}
    \dot R_{l, l'} &\;=\; \frac{\eta}{k_s}\,\sum_{p, q = 0}^{P - 1} I_3\!\big(\hat\lambda_p^{\mathrm{CNN}}, \mu_{\modP{p + l}}^{(l')}, \lambda_q\big)
    \;-\; \frac{\eta}{k_s}\,\sum_{p, q = 0}^{P - 1} I_3\!\big(\hat\lambda_p^{\mathrm{CNN}}, \mu_{\modP{p + l}}^{(l')}, \hat\lambda_q^{\mathrm{CNN}}\big) \nonumber\\
    \;&=\; \frac{P \eta}{k_s} \,\sum_{q = 0}^{P - 1} \left[I_3\!\big(\hat\lambda_p^{\mathrm{CNN}}, \mu_{\modP{p + l}}^{(l')}, \lambda_q\big)
    \;-\; I_3\!\big(\hat\lambda_p^{\mathrm{CNN}}, \mu_{\modP{p + l}}^{(l')}, \hat\lambda_q^{\mathrm{CNN}}\big)\right]\,,
\end{align}
where we have used the invariance under cyclic shifts of the positions to drop the sum over $p$ since each term of that
sum is equal and $p$-independent. Similarly, we obtain
\begin{align}\label{eq:Q_reduced_A}
    \dot Q_{l, l'} \;=\;& \frac{P \eta}{k_s} \sum_{r = 0}^{P - 1} \Big[I_3\big(\hat\lambda_p^{\mathrm{CNN}}, \nu_{\modP{p + l}}^{(l')}, \lambda_r\big) - I_3\big(\hat\lambda_p^{\mathrm{CNN}}, \nu_{\modP{p + l}}^{(l')}, \hat\lambda_r^{\mathrm{CNN}}\big)\Big] \nonumber \\
    +\;& \frac{P \eta}{k_s} \sum_{r = 0}^{P - 1} \Big[I_3\big(\hat\lambda_p^{\mathrm{CNN}}, \nu_{\modP{p + l'}}^{(l)}, \lambda_r\big) - I_3\big(\hat\lambda_p^{\mathrm{CNN}}, \nu_{\modP{p + l'}}^{(l)}, \hat\lambda_r^{\mathrm{CNN}}\big)\Big] \nonumber \\
    +\;& \frac{P \eta^2}{k_s^2}  \bigg\{\sum_{r, s = 0}^{P - 1}\Big[I_4\big(\hat\lambda_p^{\mathrm{CNN}}, \hat\lambda_q^{\mathrm{CNN}}, \lambda_r, \lambda_s\big)- 2\,I_4\big(\hat\lambda_p^{\mathrm{CNN}}, \hat\lambda_q^{\mathrm{CNN}}, \lambda_r, \hat\lambda_s^{\mathrm{CNN}}\big) \nonumber \\
    &\hspace{4.2em}  + I_4\big(\hat\lambda_p^{\mathrm{CNN}}, \hat\lambda_q^{\mathrm{CNN}}, \hat\lambda_r^{\mathrm{CNN}}, \hat\lambda_s^{\mathrm{CNN}}\big)\Big] + \sigma^2 J_2\big(\hat\lambda_p^{\mathrm{CNN}}, \hat\lambda_q^{\mathrm{CNN}}\big)\bigg\} ,
\end{align}
where again $p$ is any fixed position index and $q=\modP{p+l-l'}$.

Finally, by identifying $Q_{ll'}=\tilde{Q}_{ll'}$ and $R_{ll'}=\tilde{R}_{ll'}$ and 
by comparing term by term Eq.~\eqref{eq:R_reduced_A} with Eq.~\eqref{eq:R_reduced_B} and 
Eq.~\eqref{eq:Q_reduced_A} with Eq.~\eqref{eq:Q_reduced_B}, it is easy to show that the dynamics of the CNN
and the LCN coincide up to a time rescaling by a factor $P$.

% =============================================================================
\section{Reflection symmetry of the CNN dynamics}\label{app:reflection}
% =============================================================================

In this appendix, we show that the CNN ODEs~\eqref{eq:ODEs-CNN} are equivariant under the
reflection $\rho$ of Eq.~\eqref{eq:rho}. The idea is that reversing the kernels of both the teacher
and the student is equivalent to reflecting the input positions, $r \mapsto -r$, which leaves
the input distribution unchanged. In the ODEs, this reflection appears as a relabeling of the
fields of Appendix~\ref{app:ode-A-setup}.

We write $\bar l = k_s - 1 - l$ for the student kernel entries and $\tilde l = k_t - 1 - l$ for
the teacher ones, so that
\begin{equation}
    (\rho R)_{l, l'} = R_{\bar l, \tilde l'}\,, \qquad (\rho Q)_{l, l'} = Q_{\bar l, \bar l'}\,.
\end{equation}
We assume
that $S$ is invariant under the reflection of positions,
$S_{\modP{-r}, \modP{-q}} = S_{r, q}$, which is the case for the compound-symmetric
form~\eqref{eq:S-compound}. The teacher overlaps $T_{l, l'} = (T/k_t)\,\delta_{l, l'}$ satisfy
$T_{\tilde l, \tilde l'} = T_{l, l'}$. Finally, we denote by $\langle a\, b\rangle_y$ the
covariance of two fields $a$ and $b$ when the order parameters take the value $y = (R, Q)$.

We first show that the covariances at $\rho y$ equal those at $y$ of the following relabeled fields
\begin{equation}\label{eq:relabel}
    \nu^{(l)}_r \;\to\; \nu^{(\bar l)}_{\modP{-r}}\,, \qquad
    \mu^{(l)}_r \;\to\; \mu^{(\tilde l)}_{\modP{-r}}\,.
\end{equation}
Using the two-point functions of Appendix~\ref{app:ode-A-setup} and the invariance of $S$ and
of the teacher overlaps, we find
\begin{align}
    \langle \nu^{(l)}_r\, \nu^{(l')}_q \rangle_{\rho y}
        &= S_{r, q}\, Q_{\bar l, \bar l'}
         = S_{\modP{-r}, \modP{-q}}\, Q_{\bar l, \bar l'}
         = \langle \nu^{(\bar l)}_{\modP{-r}}\, \nu^{(\bar l')}_{\modP{-q}} \rangle_{y}\,, \\
    \langle \mu^{(l)}_r\, \mu^{(l')}_q \rangle_{\rho y}
        &= S_{r, q}\, T_{l, l'}
         = S_{\modP{-r}, \modP{-q}}\, T_{\tilde l, \tilde l'}
         = \langle \mu^{(\tilde l)}_{\modP{-r}}\, \mu^{(\tilde l')}_{\modP{-q}} \rangle_{y}\,, \\
    \langle \mu^{(l)}_r\, \nu^{(l')}_q \rangle_{\rho y}
        &= S_{r, q}\, R_{\bar l', \tilde l}
         = S_{\modP{-r}, \modP{-q}}\, R_{\bar l', \tilde l}
         = \langle \mu^{(\tilde l)}_{\modP{-r}}\, \nu^{(\bar l')}_{\modP{-q}} \rangle_{y}\,.
\end{align}
The preactivations are sums of single-vector fields, and the relabeling~\eqref{eq:relabel} maps
them to preactivations at reflected positions,
\begin{equation}\label{eq:relabel-local}
    \hat\lambda_p \;\to\; \sum_{l = 0}^{k_s - 1} \nu^{(\bar l)}_{\modP{-p-l}} = \hat\lambda_{\hat p}\,,
    \qquad
    \lambda_q \;\to\; \sum_{l = 0}^{k_t - 1} \mu^{(\tilde l)}_{\modP{-q-l}} = \lambda_{\check q}\,,
\end{equation}
with $\hat p = \modP{-p-k_s+1}$ and $\check q = \modP{-q-k_t+1}$. 

We now show that the ODEs are equivariant. Each of the Gaussian
integrals $I_2$, $J_2$, $I_3$, and $I_4$ depends on its arguments only through their covariance
matrix. For the drift of $R$ in Eq.~\eqref{eq:FR-cnn}, the relabeling
maps $\mu^{(l')}_{\modP{p + l}}$ to $\mu^{(\tilde l')}_{\modP{\hat p + \bar l}}$, and we obtain
\begin{align}
\left.\big[F_R^{\rm CNN}\big]_{l, l'} \right|_{\rho y}\;&=\; \sum_{p, q = 0}^{P - 1} \left.\Big[I_3\big(\hat\lambda_p, \mu_{\modP{p + l}}^{(l')}, \lambda_q\big) - I_3\big(\hat\lambda_p, \mu_{\modP{p + l}}^{(l')}, \hat\lambda_q\big)\Big]\right|_{\rho y} \nonumber\\
&= \left. \sum_{p, q = 0}^{P - 1} \Big[I_3\big(\hat\lambda_{\hat p}, \mu^{(\tilde l')}_{\modP{\hat p + \bar l}}, \lambda_{\check q}\big)
       - I_3\big(\hat\lambda_{\hat p}, \mu^{(\tilde l')}_{\modP{\hat p + \bar l}}, \hat\lambda_{\hat q}\big)\Big]\right|_{y} \nonumber \\
&= \left. \sum_{p', q' = 0}^{P - 1} \Big[I_3\big(\hat\lambda_{p'}, \mu^{(\tilde l')}_{\modP{p' + \bar l}}, \lambda_{ q'}\big)
       - I_3\big(\hat\lambda_{ p'}, \mu^{(\tilde l')}_{\modP{p' + \bar l}}, \hat\lambda_{q'}\big)\Big]\right|_{y}
    = \left.\big[F_R^{\rm CNN}\big]_{\bar l, \tilde l'}\right|_y\,,
\end{align}
where in going from the first to the second line we used the fact that covariances at $\rho y$ equal those at $y$ under the relabeling \eqref{eq:relabel}, and in the last step we changed the summation variables to $p' = \hat{p}$ and $q' = \check{q}$ in the first term and $q' = \hat{q}$ in the second. Analogous arguments apply to the ODE for $Q$. Writing the ODEs \eqref{eq:ODEs-CNN} in the compact form $d y /d \alpha =F(y)$, we then have the equivariance relation
\begin{equation}\label{eq:equivariance}
    F(\rho y) \;=\; \rho\, F(y)\,.
\end{equation}
Similarly, we obtain $\varepsilon_g(\rho y) =
\varepsilon_g(y)$.

Finally, let $y(\alpha)$ be the solution of Eq.~\eqref{eq:ODEs-CNN} with a $\rho$-symmetric
initial condition, $\rho\, y(0) = y(0)$. Since $\rho$ is linear, $z(\alpha) = \rho\, y(\alpha)$
satisfies, by Eq.~\eqref{eq:equivariance},
\begin{equation}
    \frac{dz}{d\alpha} \;=\; \rho\, F\big(y(\alpha)\big) \;=\; F\big(z(\alpha)\big)\,, \qquad z(0) = y(0)\,.
\end{equation}
The right-hand side $F$ is smooth, so the solution is unique and $z(\alpha) = y(\alpha)$, that
is, $\rho\, y(\alpha) = y(\alpha)$ for all $\alpha \geq 0$. The fixed-point subspace
$\mathrm{Fix}(\rho) = \{y : \rho y = y\}$ is therefore invariant under the flow.

% =============================================================================
\section{Plateau phase}\label{app:spec-checks}
% =============================================================================

\begin{figure}[t]
\centering
\begin{subfigure}{0.48\textwidth}\centering
  \includegraphics[width=\textwidth]{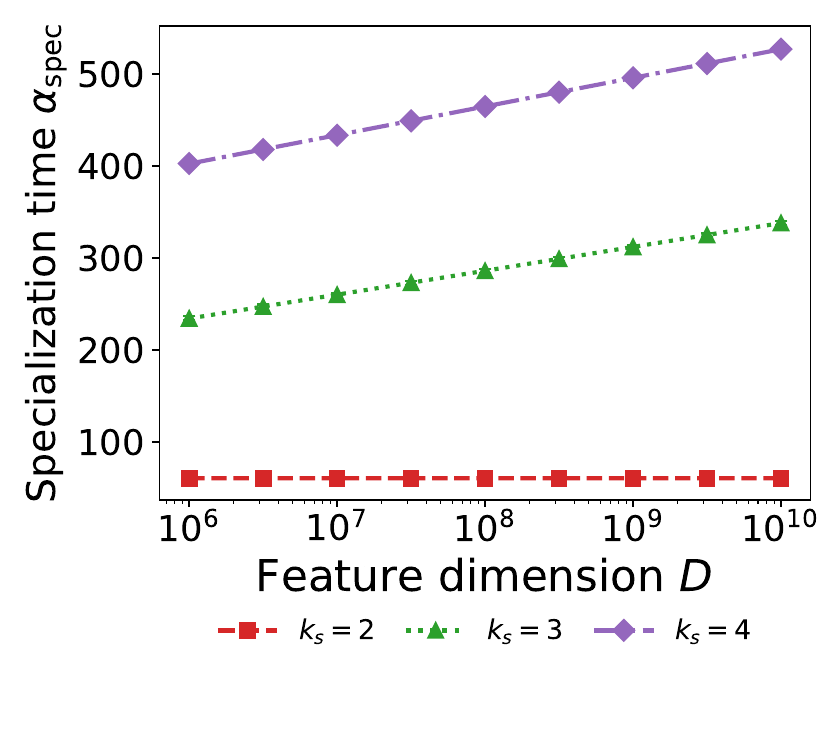}
  \caption{}\label{fig:spec-a}
\end{subfigure}
\hfill
\begin{subfigure}{0.48\textwidth}\centering
  \includegraphics[width=\textwidth]{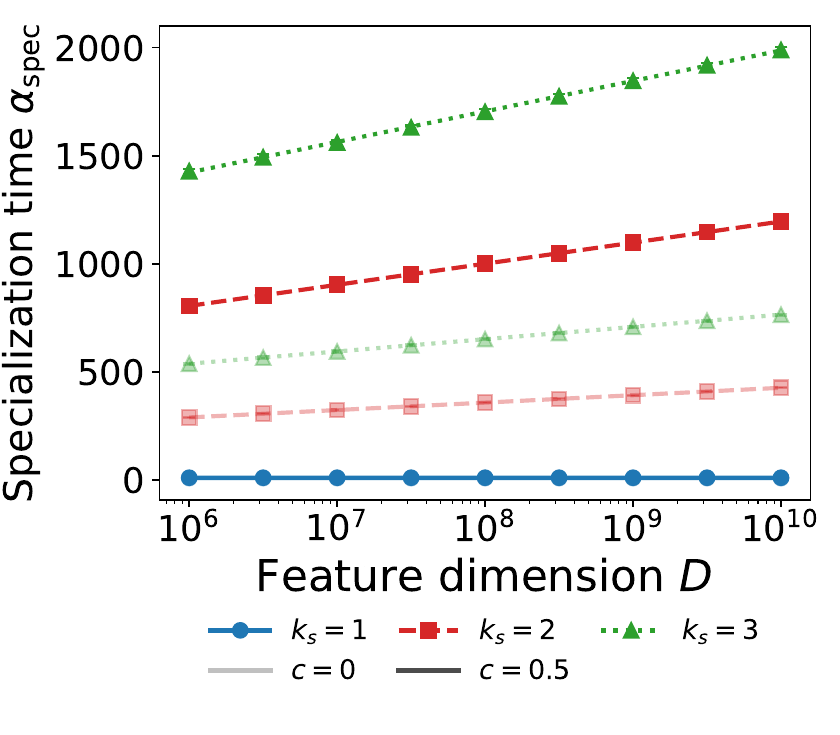}
  \caption{}\label{fig:spec-b}
\end{subfigure}
\caption{Specialization time of the CNN against the feature dimension, as in Figure~\ref{fig:plateau-b}.
Color denotes the student kernel size. Markers are averages over random initializations, each
obtained by integrating the ODEs~\eqref{eq:ODEs-CNN} from the order parameters of a random
initialization at dimension $\D$. Standard errors are smaller than the marker size.
Lines are fits $a + b\ln\D$. (a) A teacher of kernel size $k_t = 2$.
The matched case $k_s = k_t = 2$ is flat in $\D$ and the two wider
kernels grow logarithmically. (b) Correlated positions, at $k_t = 1$,
with $c = 0$ faded and $c = 0.5$ at full strength. The plateau at $k_s > k_t$ survives the
correlation.
Parameters: $g = \erf$, $P = 6$, $\eta = 0.5$ in panel (a) and $\eta = 0.2$ in panel (b),
$T = 1$, $\sigma = 0$, $c = 0$ in panel (a), $\D$ from $10^6$ to $10^{10}$, 128 random
initial conditions.}
\label{fig:spec-checks}
\end{figure}

Figure~\ref{fig:spec-checks} repeats the measurement of Figure~\ref{fig:plateau-b} in two
further settings. Panel (a) uses a teacher of kernel size $k_t = 2$. The specialization time is flat in
$\D$ at $k_s = 2$ and grows logarithmically at $k_s = 3$ and $k_s = 4$. Panel (b) repeats the
$k_t = 1$ measurement at $c = 0.5$, with $\eta = 0.2$. The specialization time grows with $\D$ for $k_s>k_t$, while it is $D$-independent for $k_s=k_t$.

% =============================================================================
\section{Early-time dynamics}\label{app:early}
% =============================================================================

The rescaling of Appendix~\ref{app:rescaling} is exact only at $S = I$. To see how the two
students differ at $c > 0$, we derive Result~\ref{res:early} by expanding the ODEs around the
random initial condition. We use the notation of Appendices~\ref{app:ode-A}
and~\ref{app:ode-B} and work at leading order in $\D$, where the initial order parameters are
$R = 0$ and $Q = I/k_s$ for the CNN, and $R^{(p)} = 0$ and $Q^{(p,q)} = \delta_{p,q}\,I/k_s$ for
the LCN. Since $S$ is compound symmetric, each entry $S_{a,b}$ equals $1$ if the positions $a$
and $b$ coincide and $c$ otherwise, so that all sums over positions below reduce to counting
coincident positions. We write $s_c \equiv \sum_q S_{p,q} = 1 + (P - 1)c$ and
$\omega(\chi) \equiv (4 - \chi^2)^{-1/2}$.

\paragraph{Initial state.}
Inserting the initial order parameters into Eqs.~\eqref{eq:covariances_eff}
and~\eqref{eq:covariances_eff_LCN} gives
\begin{equation}\label{eq:early-cov}
    \langle\lambda_p\lambda_q\rangle = T S_{p,q}, \qquad
    \langle\hat\lambda_p\lambda_q\rangle = 0, \qquad
    \langle\hat\lambda_p\hat\lambda_q\rangle =
    \begin{cases} S_{p,q} & \text{CNN},\\ \delta_{p,q} & \text{LCN}. \end{cases}
\end{equation}
The student preactivations have unit variance, and we denote by $\chi$ their correlation at
different positions, with $\chi = c$ for the CNN and $\chi = 0$ for the LCN.
The initial error follows from Eq.~\eqref{eq:gen-error-I2} of Appendix~\ref{app:gen-error},
\begin{equation}
    \varepsilon_g \;=\; \underbrace{\sum_{p, q = 0}^{P - 1} I_2\big(\lambda_p, \lambda_q\big)}_{\text{teacher}}
    \;-\; 2\,\underbrace{\sum_{p, r = 0}^{P - 1} I_2\big(\lambda_p, \hat\lambda_r\big)}_{\text{cross}}
    \;+\; \underbrace{\sum_{r, s = 0}^{P - 1} I_2\big(\hat\lambda_r, \hat\lambda_s\big)}_{\text{student}},
\end{equation}
with $I_2 = \frac{2}{\pi}\arcsin\big[c_{12}/\sqrt{(1 + c_{11})(1 + c_{22})}\big]$. Splitting each sum
into its $P$ diagonal and $P(P - 1)$ off-diagonal terms and using Eq.~\eqref{eq:early-cov}, the
teacher sum gives the mean squared output of the teacher,
\begin{equation}\label{eq:early-epsT}
    \varepsilon_T \;=\; \E\big[f^\star(x)^2\big]
    \;=\; \frac{2P}{\pi}\left[\arcsin\frac{T}{1 + T} + (P - 1)\arcsin\frac{Tc}{1 + T}\right],
\end{equation}
the cross sum vanishes, and adding the student sum gives
\begin{equation}
    \varepsilon_0 \;=\; \varepsilon_T + \frac{P}{3} + \frac{2P(P - 1)}{\pi}\arcsin\frac{\chi}{2},
\end{equation}
which yields the initial errors~\eqref{eq:early-eps0} with $\chi = 0$ for the LCN and $\chi = c$
for the CNN.

\paragraph{Equations of motion at the initial state.}
We evaluate the ODEs of Appendices~\ref{app:ode-A} and~\ref{app:ode-B} at the initial state,
where the integrals $I_3$ and $I_4$ reduce to elementary functions of $T$ and $c$, and group
the terms. For all $l < k_s$ and $l' < k_t$, the overlaps with the teacher grow as
\begin{equation}\label{eq:early-R}
    \frac{dR_{l,l'}}{d\alpha}\bigg|_{\mathrm{CNN}} = \frac{2\eta T}{\pi k_s k_t}\,\frac{P s_c}{\sqrt{2(1 + T)}} , \qquad
    \frac{dR^{(p)}_{l,l'}}{d\alpha}\bigg|_{\mathrm{LCN}} = \frac{2\eta T}{\pi k_s k_t}\,\frac{s_c}{\sqrt{2(1 + T)}} ,
\end{equation}
and the student overlaps evolve as
\begin{align}
    \frac{dQ_{l,l'}}{d\alpha}\bigg|_{\mathrm{CNN}} &= -\frac{2\eta}{k_s^2}\,H_{l,l'}
    + \frac{\eta^2}{k_s^2}\sum_{p,q = 0}^{P - 1} S_{\modP{p + l},\,\modP{q + l'}}\,V_{p,q} , \label{eq:early-Q-cnn}\\
    \frac{dQ^{(p,q)}_{l,l'}}{d\alpha}\bigg|_{\mathrm{LCN}} &= \frac{1}{k_s^2}
    \Big[-2\eta\,\Psi_{p,q} + \eta^2\,V_{p,q}\Big]\,S_{\modP{p + l},\,\modP{q + l'}} . \label{eq:early-Q-lcn}
\end{align}
For the CNN, $H_{l,l} = H_{\rm d}$ and $H_{l,l'} = H_{\rm o}$ for $l \neq l'$, with
\begin{equation}\label{eq:early-G}
    H_{\rm d} = \frac{P}{\pi}\left[\frac{1}{\sqrt 3} + (P - 1)c\,\omega(c)\right], \qquad
    H_{\rm o} = \frac{P}{\pi}\left[\frac{c}{\sqrt 3}
      + \big(2 + 2(P - 2)c - (P - 1)c^2\big)\,\omega(c)\right],
\end{equation}
and for the LCN, $\Psi_{p,p} = 1/(\pi\sqrt 3)$ and $\Psi_{p,q} = 1/\pi$ for $p \neq q$. In both cases, $V_{p,q} \equiv \langle\delta^2 g'(\hat\lambda_p)g'(\hat\lambda_q)\rangle$ is the average
in the $\mathcal{O}(\eta^2)$ term, with $\delta$ the residual of Appendix~\ref{app:ode-A}. At the
initial state it equals $V_{\rm d}$ for $p = q$ and $V_{\rm o}$ for $p \neq q$, with
\begin{align}
    V_{\rm d} &= \frac{2\big(\varepsilon_T + \sigma^2\big)}{\pi\sqrt 3}
    + \frac{4}{\pi^2\sqrt 3}\bigg[\arcsin\frac14 + 2(P - 1)\arcsin\frac{\chi}{2\sqrt{2(3 - \chi^2)}}
    + (P - 1)\arcsin\frac{3 - 2\chi^2}{2(3 - \chi^2)} \nonumber\\
    &\hspace{9em} + (P - 1)(P - 2)\arcsin\frac{\chi(3 - 2\chi)}{2(3 - \chi^2)}\bigg], \label{eq:early-V}\\
    V_{\rm o} &= \frac{2\,\omega(\chi)\big(\varepsilon_T + \sigma^2\big)}{\pi}
    + \frac{4\,\omega(\chi)}{\pi^2}\bigg[2\arcsin\frac{2 - \chi^2}{2(3 - \chi^2)} + 2\arcsin\frac{\chi}{2(3 - \chi^2)}
    \nonumber\\
    &\hspace{9em} + (P - 2)\arcsin\frac{2 + \chi - 2\chi^2}{2(2 - \chi)(1 + \chi)}
    + 4(P - 2)\arcsin\frac{\chi}{2\sqrt{(3 - \chi^2)(1 + \chi)}} \nonumber\\
    &\hspace{9em} + (P - 2)(P - 3)\arcsin\frac{\chi}{2(1 + \chi)}\bigg]. \nonumber
\end{align}
The first term comes from the teacher output and the label noise, which are independent of the
student preactivations at the initial state, and the second from the student output.

\paragraph{Expansion of the generalization error.}
Since the right-hand sides of Eqs.~\eqref{eq:early-R}--\eqref{eq:early-Q-lcn} are constant at
this order, the order parameters change linearly at small $\alpha$, and
\begin{equation}
    \frac{d\varepsilon_g}{d\alpha}\bigg|_{\alpha = 0} = \sum \frac{\partial\varepsilon_g}{\partial R}\,\frac{dR}{d\alpha}
    + \sum \frac{\partial\varepsilon_g}{\partial Q}\,\frac{dQ}{d\alpha} = -\eta A + \eta^2 B ,
\end{equation}
with the sums running over all entries of the order parameters. Differentiating
Eq.~\eqref{eq:gen-error-I2} through Eqs.~\eqref{eq:covariances_eff}
and~\eqref{eq:covariances_eff_LCN} gives the derivatives at the initial state,
\begin{align}
    \frac{\partial\varepsilon_g}{\partial R_{l,l'}}\bigg|_{\mathrm{CNN}} &= -\frac{4 P s_c}{\pi\sqrt{2(1 + T)}}, &
    \frac{\partial\varepsilon_g}{\partial Q_{l,l'}}\bigg|_{\mathrm{CNN}} &= H_{l,l'}, \label{eq:early-dE-cnn}\\
    \frac{\partial\varepsilon_g}{\partial R^{(p)}_{l,l'}}\bigg|_{\mathrm{LCN}} &= -\frac{4 s_c}{\pi\sqrt{2(1 + T)}}, &
    \frac{\partial\varepsilon_g}{\partial Q^{(p,q)}_{l,l'}}\bigg|_{\mathrm{LCN}} &= \Psi_{p,q}\,S_{\modP{p + l},\,\modP{q + l'}} . \label{eq:early-dE-lcn}
\end{align}
Inserting the equations of motion, the CNN gives
\begin{align}
    A_{\rm CNN} &= \frac{4TP^2 s_c^2}{\pi^2(1 + T)}
      + 2\left[\frac{H_{\rm d}^2}{k_s} + \Big(1 - \frac{1}{k_s}\Big)H_{\rm o}^2\right], \label{eq:early-A-cnn}\\
    B_{\rm CNN} &= P\left\{V_{\rm d}\left[\frac{H_{\rm d}}{k_s} + \Big(1 - \frac{1}{k_s}\Big)c\,H_{\rm o}\right]
    + V_{\rm o}\left[\frac{(P - 1)c\,H_{\rm d}}{k_s}
    + \Big(1 - \frac{1}{k_s}\Big)\big(1 + (P - 2)c\big)H_{\rm o}\right]\right\}, \label{eq:early-B-cnn}
\end{align}
with $V_{\rm d}$ and $V_{\rm o}$ evaluated at $\chi = c$, and the LCN gives
\begin{equation}\label{eq:early-A-lcn}
    A_{\rm LCN} = \frac{4TPs_c^2}{\pi^2(1 + T)} + \frac{2P}{\pi^2}\left[\frac{n_{\rm d}}{3} + n_{\rm o}\right], \qquad
    B_{\rm LCN} = \frac{P}{\pi}\left[\frac{n_{\rm d}}{\sqrt 3}\,V_{\rm d} + n_{\rm o}\,V_{\rm o}\right],
\end{equation}
with $V_{\rm d}$ and $V_{\rm o}$ evaluated at $\chi = 0$ and
\begin{equation}\label{eq:early-n}
    n_{\rm d} = c^2 + \frac{1 - c^2}{k_s}, \qquad
    n_{\rm o} = (1 - c^2)\Big(1 - \frac{1}{k_s}\Big) + (P - 1)c^2 .
\end{equation}
All four coefficients are positive, since every term is. They do not depend on $k_t$, and they
are affine functions of $1/k_s$. At $c = 0$ one has $A_{\rm CNN} = P A_{\rm LCN}$ and
$B_{\rm CNN} = P B_{\rm LCN}$, as required by
Result~\ref{res:rescaling}.

\paragraph{Optimal learning rate and boundary.}
Within the expansion, the error of each student initially decreases if and only if
$\eta < A/B$. At a training time $\alpha$, minimizing over $\eta$ gives $\eta^* = A/(2B)$ and
\begin{equation}
    \varepsilon^*(\alpha) \;=\; \varepsilon_0 - \alpha\,\frac{A^2}{4B} .
\end{equation}
The CNN has the lower error when
\begin{equation}\label{eq:early-crossing}
    \alpha\left[\frac{A_{\rm CNN}^2}{4B_{\rm CNN}} - \frac{A_{\rm LCN}^2}{4B_{\rm LCN}}\right]
    \;>\; \varepsilon_0^{\rm CNN} - \varepsilon_0^{\rm LCN} \;=\; \frac{2P(P - 1)}{\pi}\arcsin\frac{c}{2} .
\end{equation}
The right-hand side is positive for $c > 0$, and we find the bracket positive for all
$c \in [0, 1)$, checked numerically for $P \le 12$, all $k_s \le P$, $0.25 \le T \le 10$, and
$0 \le \sigma \le 4$. Hence, within this expansion, the LCN has the lower error at large
correlations (short training times) and the CNN at small correlations (long training times),
with a transition at the correlation $c^*(\alpha)$ where the two sides of
Eq.~\eqref{eq:early-crossing} are equal,
\begin{equation}\label{eq:early-cstar}
    c^*(\alpha) \;=\; 2\sin\!\left[\frac{\pi\alpha}{2P(P - 1)}\,\Delta\big(c^*(\alpha)\big)\right],
    \qquad
    \Delta(c) \;\equiv\; \frac{A_{\rm CNN}^2}{4B_{\rm CNN}} - \frac{A_{\rm LCN}^2}{4B_{\rm LCN}} ,
\end{equation}
an implicit equation, since $A$ and $B$ depend on $c$. At small $\alpha$, the transition occurs
at small $c^*$, and expanding Eq.~\eqref{eq:early-cstar} to first order in $\alpha$, with $A$ and $B$
evaluated at $c = 0$, gives $c^*(\alpha) = c_0\,\alpha + \mathcal{O}(\alpha^2)$, with
\begin{equation}\label{eq:c0}
    c_0 \;=\; \frac{1}{\pi\beta}\left(1 + \frac{2T}{1 + T} - \frac{2}{3k_s}\right)^2 ,
\end{equation}
where
\begin{equation}\label{eq:c0-beta}
    \beta \;=\; \frac{2}{3k_s}\left[\bar\varepsilon + \frac{P - 1}{3} + \frac{2}{\pi}\arcsin\frac14\right]
    + \Big(1 - \frac{1}{k_s}\Big)\left[\bar\varepsilon + \frac{P - 2}{3} + \frac{4}{\pi}\arcsin\frac13\right],
\end{equation}
and $\bar\varepsilon = \frac{2P}{\pi}\arcsin\frac{T}{1 + T} + \sigma^2$.

% =============================================================================
\section{Noise-floor calculation}\label{app:perturbative}
% =============================================================================

In this appendix, we investigate the long-time behavior where the network reaches near-perfect recovery.
For $\eta\to0$ perfect recovery is obtained asymptotically and the student exactly reproduces the teacher.
At finite but small $\eta$, we perform a perturbative calculation to locate the 
fixed point under label noise and the
corresponding generalization error, for the CNN and the LCN and for any $k_t \le k_s \le P$. Throughout
this appendix $l, l', l''$ label taps and $p, q, r, s$ label positions.

\subsection{CNN student}
Write $u_l = \hat w_l - \sqrt{k_s/k_t}\,w_l$, with the zero-extended teacher kernel of
Appendix~\ref{app:ode-A-setup}, for the deviation of the student's kernel from
perfect recovery, and define the single-tap error field
\begin{equation}\label{eq:eps-field}
    \epsilon^{(l)}_p \;\equiv\; \nu^{(l)}_p - \mu^{(l)}_p \;=\; \frac{u_l \cdot x_p}{\sqrt{k_s\D}} .
\end{equation}
Its covariance follows from the two-point functions of Appendix~\ref{app:ode-A-setup},
\begin{equation}\label{eq:eps-cov}
    \big\langle \epsilon^{(l)}_p\,\epsilon^{(l')}_q \big\rangle \;=\; S_{p, q}\,M_{l, l'},
    \qquad
    M \;\equiv\; Q - R - R^{\top} + T ,
\end{equation}
where $M$ is a symmetric $k_s \times k_s$ matrix that vanishes exactly at perfect recovery.
The error in the preactivation at position $p$ is
\begin{equation}\label{eq:e-field}
    e_p \;\equiv\; \hat\lambda_p - \lambda_p \;=\; \sum_{l = 0}^{k_s-1} \epsilon^{(l)}_{\modP{p + l}} ,
\end{equation}
so that its covariance is the effective matrix built from $M$,
\begin{equation}\label{eq:e-cov}
    \langle e_p e_q \rangle \;=\; M^{\Eff}_{p, q} \;=\;
    \sum_{l, l' = 0}^{k_s - 1} M_{l, l'}\,S_{\modP{p + l},\,\modP{q + l'}} .
\end{equation}
Near perfect recovery and for small learning rate $\eta$, we will find that $M$ is of order $\eta$, the fields
$e_p$ are then of order $\sqrt\eta$, and the correlations with the teacher fields,
$\langle e_p \lambda_q \rangle = (R - T)^{\Eff}_{p,q}$, are of order $\eta$. 

Expanding the residual for small field errors,
\begin{equation}\label{eq:Dy-expand}
    \Delta y \;=\; -\sum_{p = 0}^{P-1}\Big[g(\lambda_p + e_p) - g(\lambda_p)\Big]
      \;=\; -\sum_{p = 0}^{P-1} g'(\lambda_p)\,e_p \;+\; \mathcal{O}(e^2) ,
\end{equation}
so that, writing $W_{p, q} \equiv \langle g'(\lambda_p) g'(\lambda_q)\rangle$,
\begin{equation}\label{eq:eps-WM}
    \varepsilon^{\infty} \;=\; \big\langle (\Delta y)^2 \big\rangle
      \;=\; \sum_{p, q = 0}^{P-1} W_{p, q}\,\langle e_p e_q\rangle \;+\; \mathcal{O}(\eta^2)
      \;=\; \sum_{p, q = 0}^{P-1} W_{p, q}\,M^{\Eff}_{p, q} \;+\; \mathcal{O}(\eta^2) \,,
\end{equation}
where we have used
\begin{equation}
    \langle g'(\lambda_p)g'(\lambda_q)\,e_p e_q\rangle = W_{p,q}\langle e_p
    e_q\rangle +\mathcal{O}(\eta^2)\,.
\end{equation}
Using Eq.~\eqref{eq:J2def}, we find
\begin{equation}\label{eq:W-entries}
    W_{p, p} = \frac{2}{\pi\sqrt{1 + 2T}},
    \qquad
    W_{p, q} = \frac{2}{\pi\sqrt{(1 + T)^2 - T^2 c^2}} \quad (p \neq q) .
\end{equation}
It is convenient to define the $k_s \times k_s$ matrix
\begin{equation}\label{eq:Omega-def}
    \Omega_{l, l'} \;\equiv\; \sum_{p, q = 0}^{P-1} W_{p, q}\,S_{\modP{p + l},\,\modP{q + l'}} ,
\end{equation}
so that substituting \eqref{eq:e-cov} into \eqref{eq:eps-WM} gives
\begin{equation}\label{eq:eps-trace}
    \varepsilon^{\infty} \;=\; \sum_{l, l' = 0}^{k_s-1} M_{l, l'}\,\Omega_{l, l'} \;=\; \Tr\big(M\,\Omega\big) .
\end{equation}

We now need the equation of motion for $M$. Using the definition in
\eqref{eq:eps-cov} and the ODEs of Appendix~\ref{app:ode-A}, we find
\begin{equation}\label{eq:Mdot-raw}
    \dot M_{l, l'} \;=\; \frac{\eta}{k_s} \sum_{p = 0}^{P-1}
      \big\langle g'(\hat\lambda_p)\,\epsilon^{(l')}_{\modP{p + l}}\,\Delta y\big\rangle
      \;+\; (l \leftrightarrow l')
      \;+\; \dot Q_{l, l'}\big|_{\mathcal{O}(\eta^2)} .
\end{equation}
In the first term, we can replace $g'(\hat\lambda_p)$ by $g'(\lambda_p)$ to leading order. Using \eqref{eq:Dy-expand}, we find
\begin{equation}
    \big\langle g'(\hat\lambda_p)\,\epsilon^{(l')}_{\modP{p + l}}\,\Delta y\big\rangle =  \big\langle g'(\lambda_p)\,\epsilon^{(l')}_{\modP{p+l}}\,\Delta y \big\rangle
   +o(\eta) \;=\; -\sum_{q = 0}^{P-1}\sum_{l'' = 0}^{k_s-1} W_{p, q}\,
      S_{\modP{p+l},\,\modP{q+l''}}\,M_{l', l''} +o(\eta).
\end{equation}
Using Eq.~\eqref{eq:dQ-full} and keeping terms up to quadratic order in $\eta$ (recall that $M=\mathcal{O}(\eta)$), we get
\begin{equation}\label{eq:Mdot}
    \dot M \;=\; -\frac{\eta}{k_s}\,\big(\Omega M + M \Omega\big)
      \;+\; \frac{\eta^2\sigma^2}{k_s^2}\, \Omega +o(\eta^2) .
\end{equation}

Setting $\dot M = 0$ to leading order, we find the
Lyapunov equation
\begin{equation}\label{eq:lyapunov}
    \Omega M + M \Omega \;=\; \frac{\eta\,\sigma^2}{k_s}\,\Omega \,.
\end{equation}
It can be checked that $\Omega$ is positive definite for $c<1$. Therefore, Eq.~\eqref{eq:lyapunov} has the unique solution
\begin{equation}\label{eq:M-sol}
    M_{l,l'} \;=\; \frac{\eta\sigma^2}{2k_s}\;\delta_{l,l'} \,.
\end{equation}
Substituting \eqref{eq:M-sol} into \eqref{eq:eps-trace},
\begin{equation}\label{eq:floor-trace}
    \varepsilon^{\infty} \;=\; \frac{\eta\sigma^2}{2k_s}\,\Tr\big(\Omega\big) .
\end{equation}
From \eqref{eq:Omega-def} and using $S_{\modP{p+l},\modP{q+l}} =
S_{p,q}$, we find
\begin{equation}\label{eq:trOmega}
    \Tr\,\Omega \;=\; \sum_{l = 0}^{k_s-1}\sum_{p, q = 0}^{P-1} W_{p, q}\,S_{\modP{p+l},\,\modP{q+l}}
    \;=\; k_s \sum_{p, q = 0}^{P-1} W_{p, q}\,S_{p, q} \,.
\end{equation}
Finally, using \eqref{eq:W-entries} we recover \eqref{eq:eps-inf-A},
\begin{equation}\label{eq:eps-inf-A-full}
    \varepsilon^{\infty,\mathrm{CNN}}(c)
    \;=\; \frac{\eta\sigma^2}{\pi}\left[\frac{P}{\sqrt{1 + 2T}}
      \;+\; \frac{P(P-1)\,c}{\sqrt{(1 + T)^2 - T^2c^2}}\right] ,
\end{equation}
with no dependence on either kernel size.

\subsection{LCN student}

The LCN calculation follows the same steps but every position
carries its own kernel, so the deviation from perfect recovery has a position label,
\begin{equation}\label{eq:u-B}
    u^{(p)}_l \;\equiv\; \hat w^{(p)}_l - \sqrt{k_s/k_t}\;w_l ,
\end{equation}
and so does the single-tap error field
\begin{equation}\label{eq:eps-field-B}
    \epsilon^{(p,l)}_r \;\equiv\; \nu^{(p,l)}_r - \mu^{(l)}_r
    \;=\; \frac{u^{(p)}_l \cdot x_r}{\sqrt{k_s\D}} .
\end{equation}
Its covariance is
\begin{equation}\label{eq:M-B-def}
    \big\langle \epsilon^{(p,l)}_r\,\epsilon^{(q,l')}_{r'} \big\rangle
      \;=\; S_{r, r'}\,M^{(p,q)}_{l, l'} ,
    \qquad
    M^{(p,q)}_{l, l'} \;\equiv\; \frac{u^{(p)}_l \cdot u^{(q)}_{l'}}{k_s\D}
      \;=\; \Big[Q^{(p,q)} - R^{(p)} - \big(R^{(q)}\big)^{\top} + T\Big]_{l, l'} ,
\end{equation}
which vanishes at perfect recovery. The error in the preactivation at position $p$ is
\begin{equation}\label{eq:e-field-B}
    e_p \;\equiv\; \hat\lambda^{\mathrm{LCN}}_p - \lambda_p
    \;=\; \sum_{l = 0}^{k_s-1} \epsilon^{(p,l)}_{\modP{p + l}} .
\end{equation}

The equation of motion for $M^{(p,q)}_{l, l'}$ reads to leading order in $\eta$
\begin{equation}\label{eq:Mdot-B}
    \dot M^{(p,q)}_{l, l'} \;=\;
      -\frac{\eta}{k_s} \sum_{r = 0}^{P-1}\sum_{l'' = 0}^{k_s-1}
        W_{p, r}\,S_{\modP{p+l},\,\modP{r+l''}}\,M^{(q, r)}_{l', l''}
      \;+\; (p \leftrightarrow q,\; l \leftrightarrow l')
      \;+\; \frac{\eta^2\sigma^2}{k_s^2}\,S_{\modP{p+l},\,\modP{q+l'}}\,W_{p, q} \,.
\end{equation}
To solve the fixed-point equation $\dot M^{(p,q)}_{l, l'}=0$, we consider the ansatz
\begin{equation}\label{eq:M-ansatz-B}
    M^{(p,q)}_{l,l'} \;=\; m\,\delta_{p,q}\,\delta_{l,l'} ,
\end{equation}
with $m$ a single scalar to be fixed. We find
\begin{equation}
    \dot M^{(p,q)}_{l, l'} \;=\;
    \left[-\frac{2\eta m}{k_s} + \frac{\eta^2\sigma^2}{k_s^2}\right]
      W_{p, q}\,S_{\modP{p+l},\,\modP{q+l'}} =0,
\end{equation}
which is solved by $m = \eta\sigma^2/(2k_s)$. Therefore, we obtain
\begin{equation}\label{eq:M-sol-B}
    M^{(p,q)}_{l,l'} \;=\; \frac{\eta\sigma^2}{2k_s}\,\delta_{p, q}\delta_{l,l'} .
\end{equation}
Substituting \eqref{eq:M-sol-B} into $\langle e_p
e_q \rangle = \sum_{l,l'} M^{(p,q)}_{l,l'} S_{\modP{p+l},\modP{q+l'}}$, we find
\begin{equation}
    \langle e_p e_q \rangle \;=\; \frac{\eta\sigma^2}{2}\,\delta_{p,q} ,
\end{equation}
which is \eqref{eq:err-fields}. Only the diagonal of \eqref{eq:eps-WM} survives, giving
\eqref{eq:eps-inf-B},
\begin{equation}\label{eq:eps-inf-B-full}
    \varepsilon^{\infty,\mathrm{LCN}}
    \;=\; \frac{\eta\sigma^2}{2} \sum_{p = 0}^{P-1} W_{p, p}
    \;=\; \frac{\eta\sigma^2\,P}{\pi\sqrt{1 + 2T}} ,
\end{equation}
independent of $c$ and of both kernel sizes.

% =============================================================================
\section{Phase diagram: numerical method}\label{app:phase}
% =============================================================================

\begin{figure}[t]
\centering
\begin{subfigure}{0.48\textwidth}\centering
  \includegraphics[width=\textwidth]{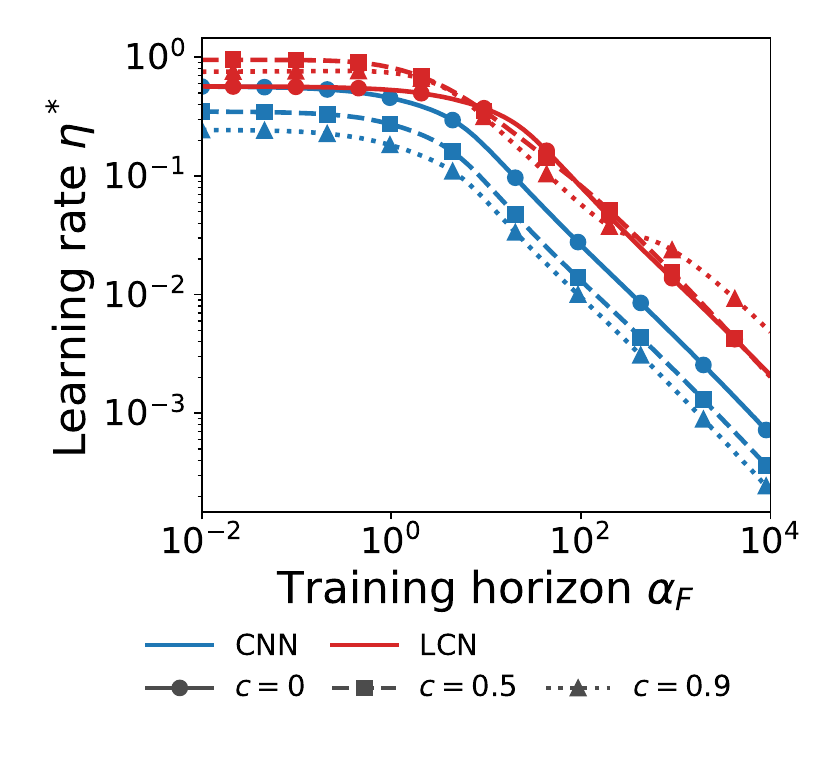}
  \caption{}\label{fig:phase-app-a}
\end{subfigure}
\hfill
\begin{subfigure}{0.48\textwidth}\centering
  \includegraphics[width=\textwidth]{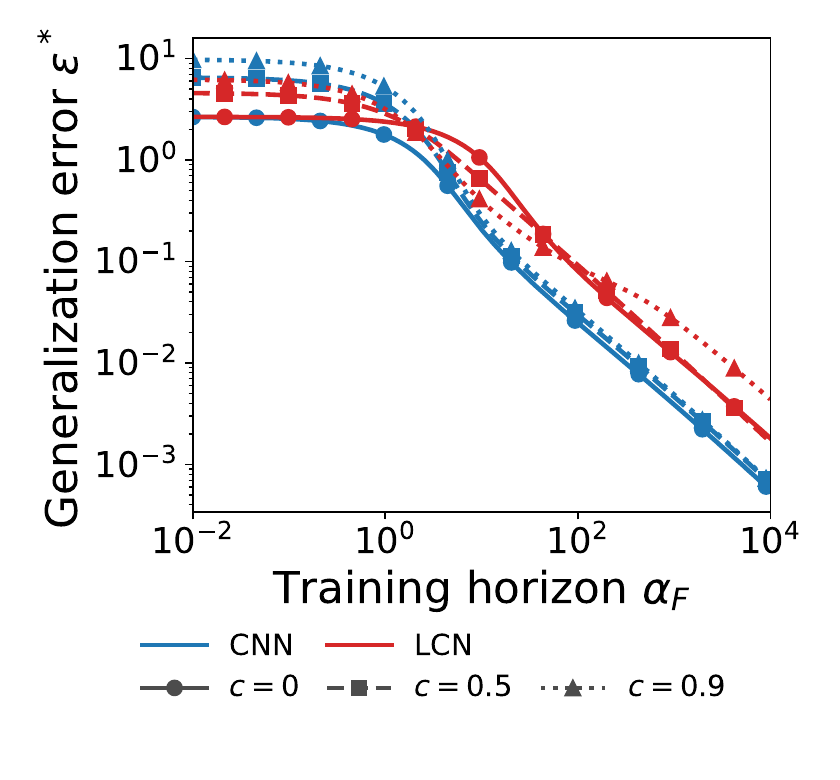}
  \caption{}\label{fig:phase-app-b}
\end{subfigure}
\caption{(a) Optimal learning rate $\eta^*$ and (b) corresponding generalization error
$\varepsilon^*$ (Section~\ref{sec:phase}) against the training horizon $\alpha_F$, for the CNN and
the LCN at three values of the correlation $c$, from the same ODE solutions as
Figure~\ref{fig:phase}. Parameters: $g = \erf$, $P = 4$, $k_s = k_t = 1$, $T = 1$, $\sigma = 1$,
random initial condition in the limit $\D \to \infty$.}
\label{fig:phase-app}
\end{figure}

For each student and each correlation $c \in \{0, 0.025, \ldots, 0.975\}$, we integrate the
ODEs~\eqref{eq:ODEs-CNN} and~\eqref{eq:ODEs-LCN} from the random initial condition at leading
order in $\D$, that is, Eqs.~\eqref{eq:init-OP-CNN} and~\eqref{eq:init-OP-LCN} without the
$\mathcal{O}_{\D}(\D^{-1/2})$ corrections. We use 80 learning rates, logarithmically spaced
between $10^{-4}$ and $3$, and record the generalization error at 400 training horizons
$\alpha_F$, logarithmically spaced between $10^{-2}$ and $10^4$.
At each $\alpha_F$, we take the learning rate with the lowest error on this grid and refine the
minimum by fitting a parabola to $\log\varepsilon_g$ as a function of $\log\eta$ through this
point and its two neighbors. The minimum never lies at the edge of the grid.
As a check, at $c = 0$ the results satisfy
$\varepsilon^*_{\rm LCN}(P\alpha_F) = \varepsilon^*_{\rm CNN}(\alpha_F)$ to within $0.4\%$, as
implied by Result~\ref{res:rescaling}.

Figure~\ref{fig:phase-app} shows the optimal learning rate and the corresponding error for
three values of $c$. For $c > 0$, the optimal learning rate of the CNN is lower than that of the
LCN at every training horizon. The CNN can afford it because it approaches the teacher faster at
small learning rates.

\begin{figure}[t]
\centering
\begin{subfigure}{0.48\textwidth}\centering
  \includegraphics[width=\textwidth]{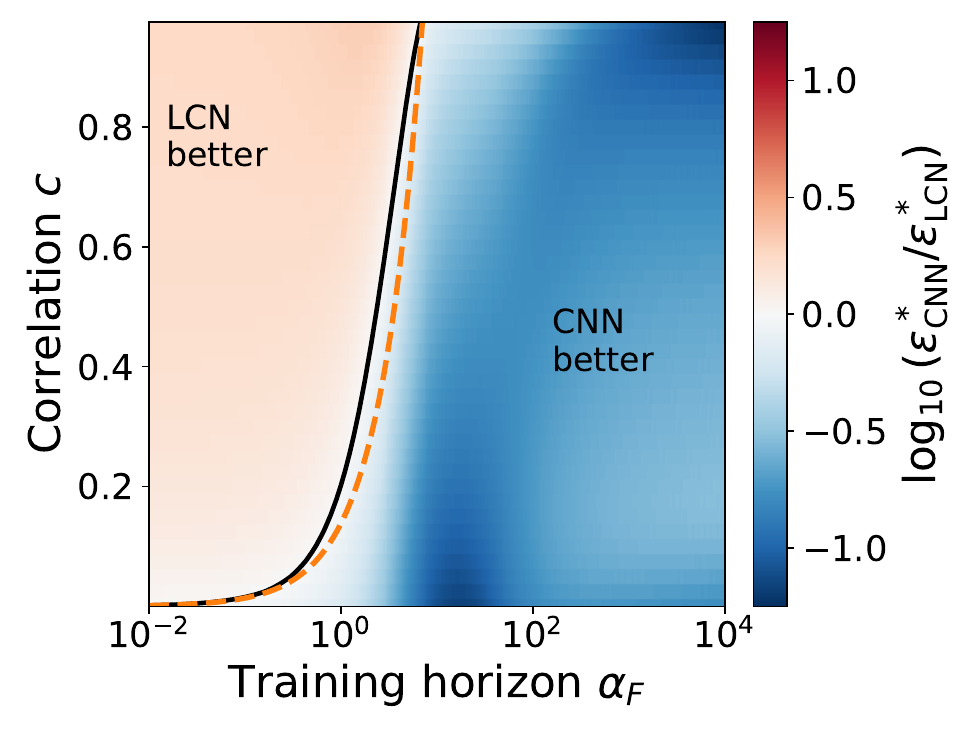}
  \caption{}\label{fig:phase-var-a}
\end{subfigure}
\hfill
\begin{subfigure}{0.48\textwidth}\centering
  \includegraphics[width=\textwidth]{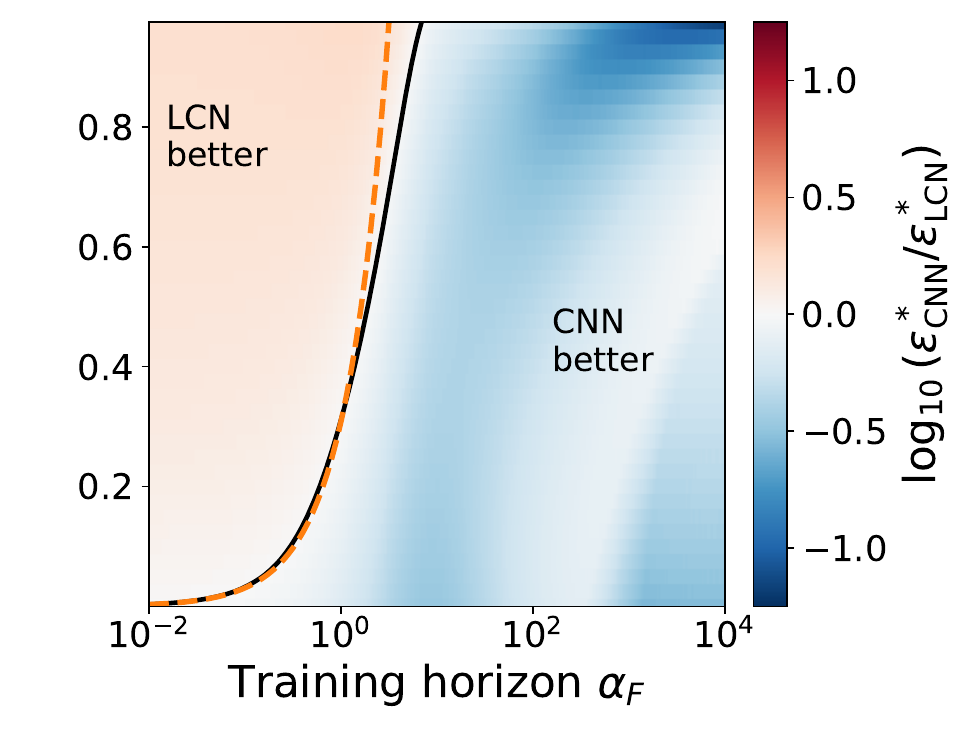}
  \caption{}\label{fig:phase-var-b}
\end{subfigure}
\begin{subfigure}{0.48\textwidth}\centering
  \includegraphics[width=\textwidth]{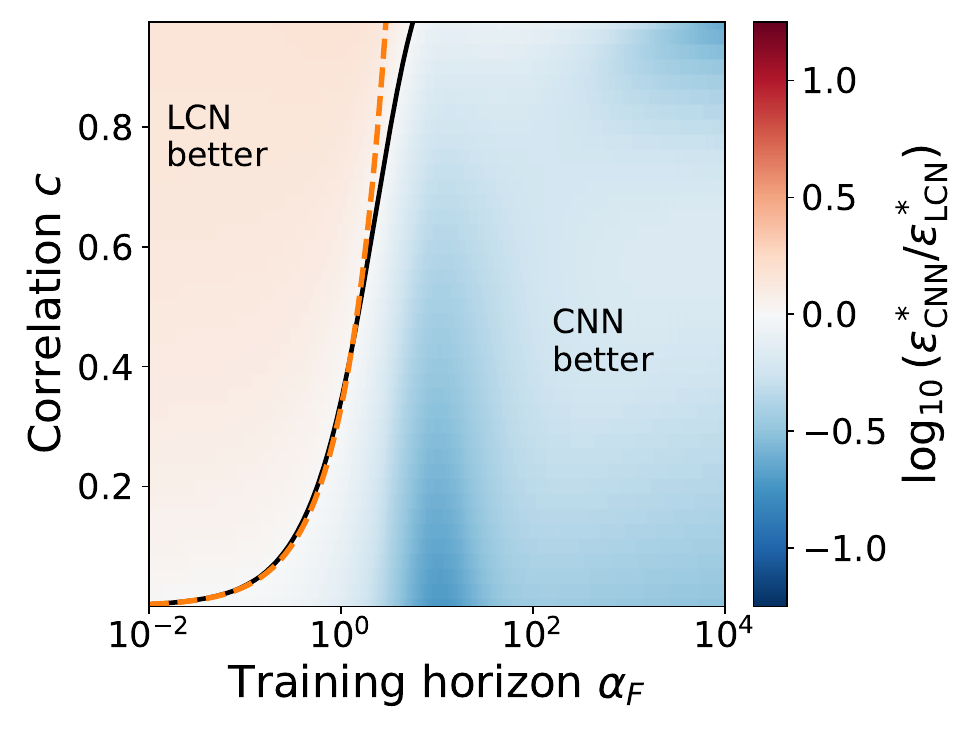}
  \caption{}\label{fig:phase-var-c}
\end{subfigure}
\hfill
\begin{subfigure}{0.48\textwidth}\centering
  \includegraphics[width=\textwidth]{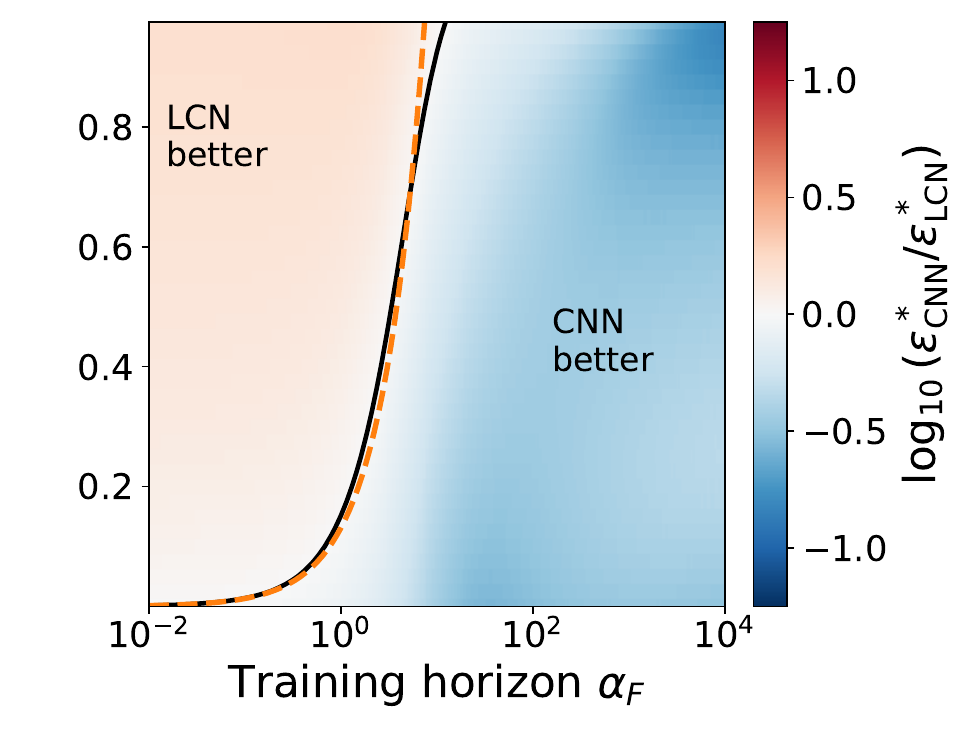}
  \caption{}\label{fig:phase-var-d}
\end{subfigure}
\caption{Phase diagram of Figure~\ref{fig:phase} for other parameter values. The color shows
$\log_{10}(\varepsilon^*_{\rm CNN}/\varepsilon^*_{\rm LCN})$ on a common scale. Solid line:
boundary $\varepsilon^*_{\rm CNN} = \varepsilon^*_{\rm LCN}$, obtained by integrating the ODEs
numerically. Dashed line: analytical prediction $c = c_0\alpha_F$ of the early-time expansion,
Eq.~\eqref{eq:cstar}. Each panel changes one
parameter of Figure~\ref{fig:phase}: (a) $P = 8$, (b) $k_s = k_t = 2$, (c) $T = 2$,
(d) $\sigma = 2$. Other parameters: $g = \erf$, $P = 4$, $k_s = k_t = 1$, $T = 1$,
$\sigma = 1$, random initial condition in the limit $\D \to \infty$.}
\label{fig:phase-var}
\end{figure}

Figure~\ref{fig:phase-var} shows the phase diagram for other parameter values, each differing
from Figure~\ref{fig:phase} in a single parameter. In all cases, for every $c > 0$, the LCN has
the lower error at short training horizons and the CNN at long ones, with a single transition.

% =============================================================================
\section{Activation dependence: ReLU}\label{app:relu}
% =============================================================================

\begin{figure}[t]
\centering
\includegraphics[width=0.6\textwidth]{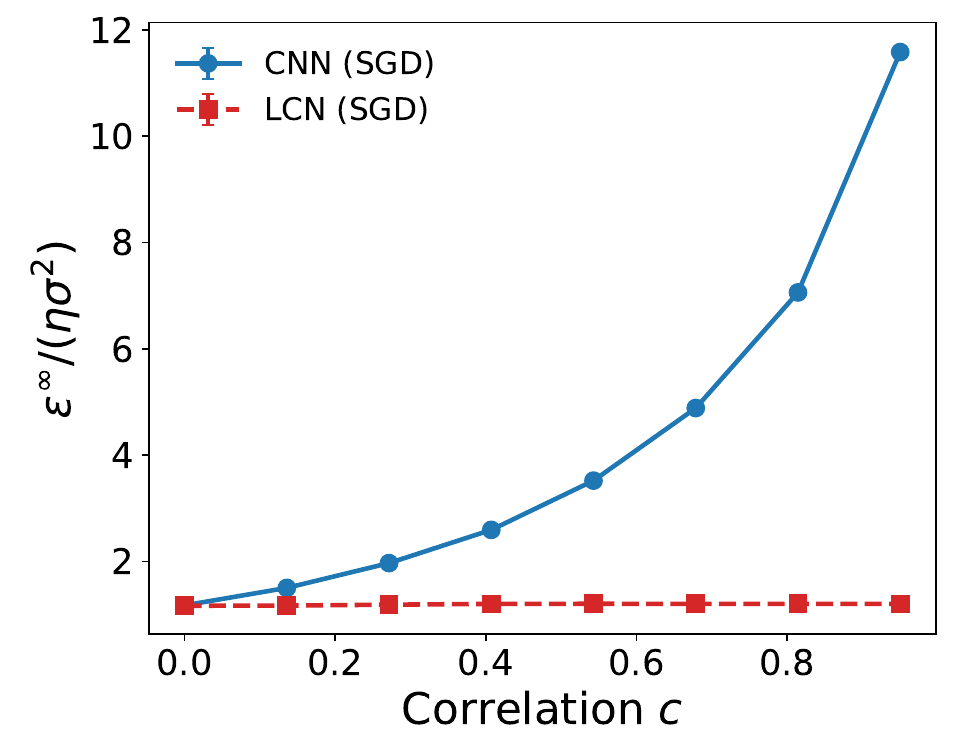}
\caption{Rescaled noise floor $\varepsilon^\infty/(\eta\sigma^2)$ of the two students with
the ReLU activation, where $\varepsilon^\infty$ is the asymptotic generalization error under
label noise. It is measured by online SGD from the specialized initial condition and
averaged over seeds, with the generalization error computed exactly from the student weights.
Standard errors are smaller than the marker size. As for $\erf$ (Figure~\ref{fig:floor-vs-c}), the CNN's floor grows with the data
correlation $c$ while the LCN's stays flat. Parameters: $g(z) = \max(0, z)$, $k_s = k_t = 2$, $P =
4$, $\D = 400$, $\eta = 0.1$, $T = 1$, $\sigma = 1$, runs up to $\alpha = 800$, $8$ seeds.}
\label{fig:relu-floor}
\end{figure}

\begin{figure}[t]
\centering
\begin{subfigure}{0.48\textwidth}\centering
  \includegraphics[width=\textwidth]{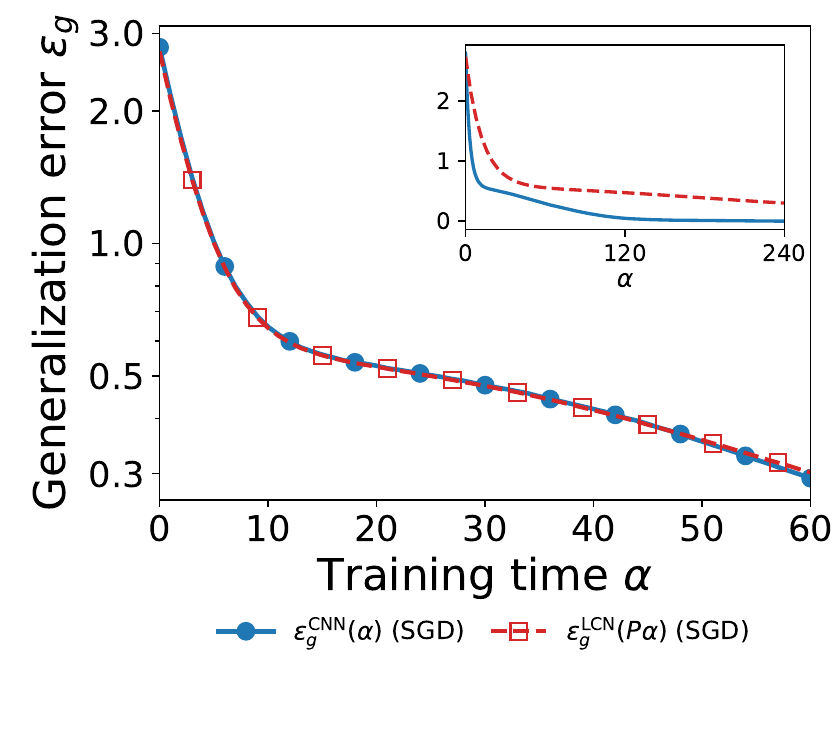}
  \caption{}\label{fig:relu-rescaling-a}
\end{subfigure}
\hfill
\begin{subfigure}{0.48\textwidth}\centering
  \includegraphics[width=\textwidth]{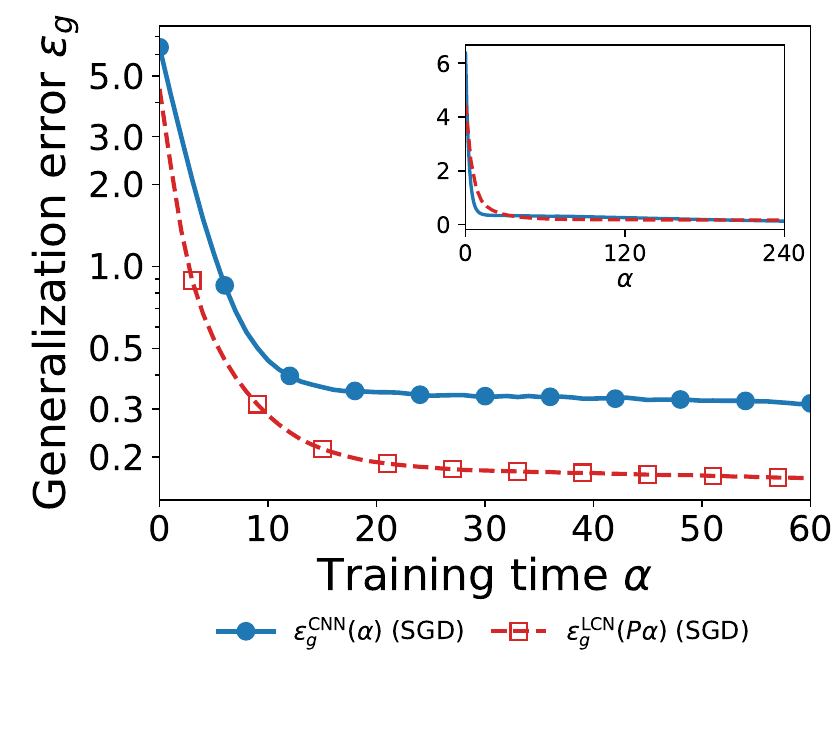}
  \caption{}\label{fig:relu-rescaling-b}
\end{subfigure}
\caption{Rescaling test of Figure~\ref{fig:rescaling} with the ReLU activation. Generalization
errors $\varepsilon_g^{\rm CNN}(\alpha)$ of the CNN and $\varepsilon_g^{\rm LCN}(P\alpha)$ of the LCN,
from online SGD averaged over 16 runs per student. Standard errors are smaller than the marker
size. Insets show the curves without the time rescaling. (a) $c = 0$.
(b) $c = 0.5$. Parameters: $g(z) = \max(0, z)$, $k_t = 2$, $k_s = 3$, $P = 4$, $\eta = 0.2$,
$\sigma = 0$, $T = 1$, $\D = 1000$.}
\label{fig:relu-rescaling}
\end{figure}

The analytic results use $g(z) = \erf(z/\sqrt 2)$ because it makes the order-parameter
equations closed form. The qualitative conclusions, however, do
not depend on this choice. Figure~\ref{fig:relu-floor} repeats the noise-floor comparison
of Figure~\ref{fig:floor-vs-c} with the ReLU activation $g(z) = \max(0, z)$. The picture is the same as for
$\erf$: the floor of the CNN grows with the data correlation $c$, while the floor of the LCN
stays flat. The mechanism, coherent accumulation of correlated SGD noise into a shared kernel,
is a property of weight sharing and not of the activation.
Likewise, the $P$-fold rescaling of Section~\ref{sec:AvsB-rescaling} holds for ReLU at $c = 0$
and fails at $c = 0.5$ (Figure~\ref{fig:relu-rescaling}).

\end{document}